\documentclass[10pt,twocolumn,letterpaper]{article}

\usepackage{cvpr}              

\usepackage{multirow}
\usepackage{array}

\definecolor{cvprblue}{rgb}{0.21,0.49,0.74}
\usepackage[pagebackref,breaklinks,colorlinks,allcolors=cvprblue]{hyperref}

\def\paperID{6153} 
\def\confName{CVPR}
\def\confYear{2026}

\title{ActiveGrasp: Information-Guided Active Grasping with Calibrated Energy-based Model}

\author{ 
Boshu Lei\textsuperscript{1}\thanks{equal contribution} \quad Wen Jiang\textsuperscript{1}\footnotemark[1] \quad  Kostas Daniilidis\textsuperscript{1,2}\\
        $^1$ University of Pennsylvania \qquad
        $^2$ Archimedes, Athena RC\\
        {\tt\small \{leiboshu, wenjiang\}@seas.upenn.edu, kostas@cis.upenn.edu}
}
\begin{document}
\maketitle
\begin{abstract}
Grasping in a densely cluttered environment is a challenging task for robots. 
Previous methods tried to solve this problem by actively gathering multiple views before grasp pose generation. 
However, they either overlooked the importance of the grasp distribution for information gain estimation or relied on the projection of the grasp distribution, which ignores the structure of grasp poses on the SE(3) manifold. 
To tackle these challenges, we propose a calibrated energy-based model for grasp pose generation and an active view selection method that estimates information gain from grasp distribution. 
Our energy-based model captures the multi-modality nature of grasp distribution on the SE(3) manifold. 
The energy level is calibrated to the success rate of grasps so that the predicted distribution aligns with the real distribution.
The next best view is selected by estimating the information gain for grasp from the calibrated distribution conditioned on the reconstructed environment, which could efficiently drive the robot to explore affordable parts of the target object. 
Experiments on simulated environments and real robot setups demonstrate that our model could successfully grasp objects in a cluttered environment with limited view budgets compared to previous state-of-the-art models. 
Our simulated environment can serve as a reproducible platform for future research on active grasping.
Website: \href{https://rpfey.github.io/activegrasp/}{https://rpfey.github.io/activegrasp/}
\end{abstract}    
\section{Introduction}
\label{sec:intro}

\begin{figure}
    \centering
    \includegraphics[width=\linewidth]{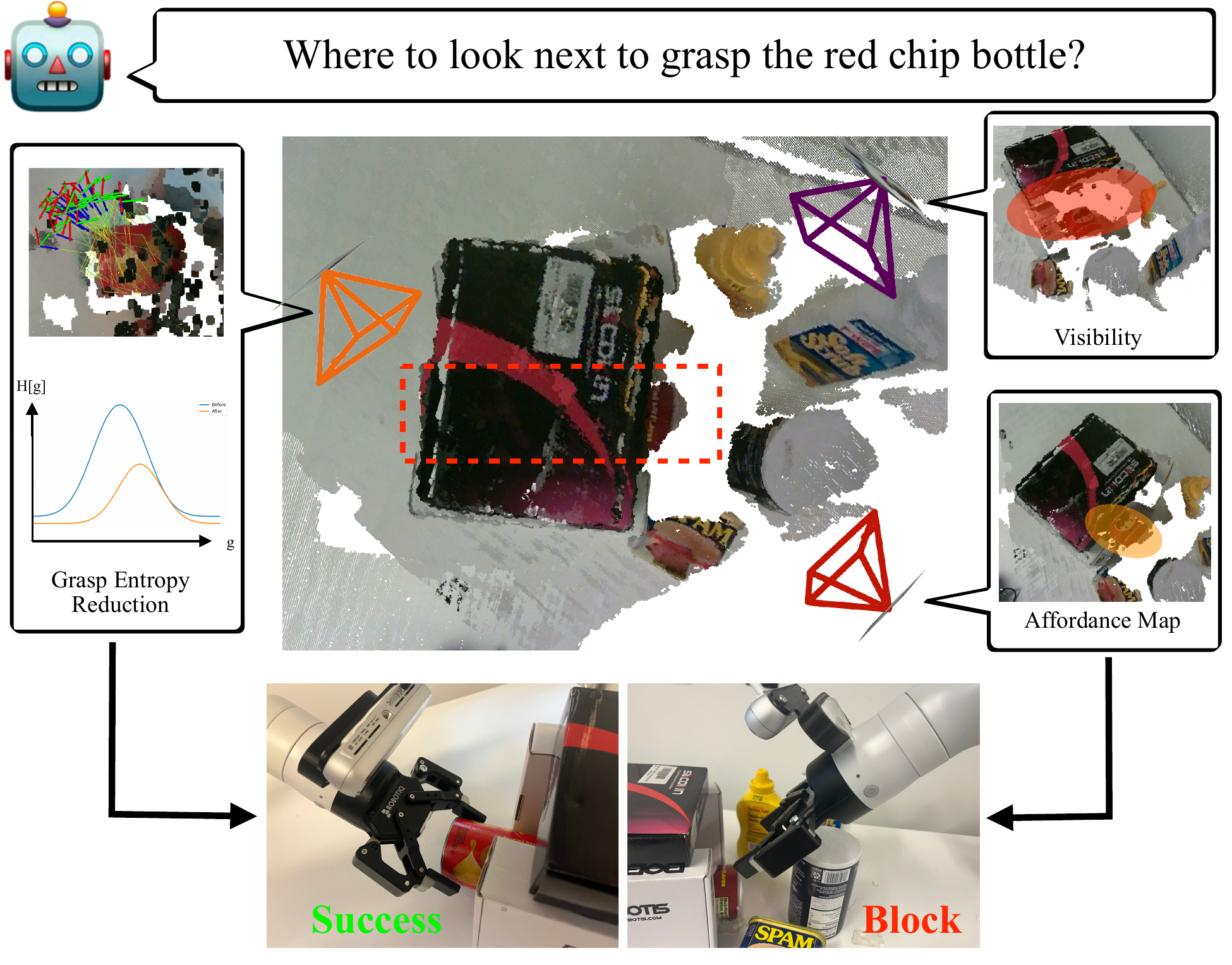}
    \caption{\textbf{ActiveGrasp Overview} The ActiveGrasp framework actively selects the next view with the highest information gain for the grasp task. Previous methods estimate the information gain using visibility or the affordance map, focusing on regions where visual features are rich but infeasible for grasp. In contrast, our method estimates information gain as entropy reduction of grasp pose distribution, selecting a region that observes highly uncertain grasps.}
    \label{fig:teaser}
    \vspace{-5mm}
\end{figure}

Grasping in clutter is a fundamental challenging task in robot manipulation, since the initial given views could not provide useful information to predict successful grasps.
Many attempts have been made to improve the performance of the model in a clutter environment by training better grasping models with larger amounts of data from real-world data and massive simulations~\cite{fang2023anygrasp, mahler2017dexnet20deeplearning,levinelearningrobotgrasping2018, robertplatthighprecisiongraspindense2016}. 
To the best of these methods, even large Vision-Language-Action (VLA) models like $\pi_{0.5}$~\cite{pi05} struggle to grasp an object in a cluttered environment due to a lack of active information gathering.
To solve this, a series of works~\cite{zhang2023ace,breyer2022closedloop,ma2024activeNGF,douglasmultiviewpicking} actively take the next-best-view (NBV) by selecting the view with the highest information gain for the grasping task before prediction. 

A key challenge for NBV selection is to estimate the information gain of candidate views for the grasping task. 
From information theory, we provide three necessary criteria for an unbiased estimate of information gain. First, the information gain should be computed from the grasp distribution instead of visibility~\cite{breyer2022closedloop,arrudaactivevisionfordexterousgraspingfornovelobjects,gregoryactiveexplorationusingtrajectoryoptimization}, which will bias towards scene completion. Second, the grasp distribution should be formulated on the SE(3) manifold instead of 2D~\cite{ma2024activeNGF,zhang2023ace} or 3D projection~\cite{douglasmultiviewpicking}, which will bias towards the position of grasps. Third, the grasp distribution should be calibrated to the real grasp distribution such that the information gain computed on top of it is unbiased.

To meet all the criteria above, we propose ActiveGrasp, a novel approach to estimate the information gain for grasping tasks by using a calibrated energy-based model to estimate the grasp distribution on the SE(3) manifold.
Rather than relying on other surrogates to estimate information gain, we define the information gain as the entropy reduction of the grasp distribution and estimate the entropy using the energy-based model, providing more task-specific and accurate information gain estimation. 

To be specific, we define the entropy of grasp poses as the expectation for the entropy of a single grasp.  
The entropy of a single grasp can be computed using the success rate, as we model every single grasp as a Bernoulli distribution.
The expectation is computed using the Gaussian Approximation of Posterior~\cite{kirsch2022unifying} (GAP) of the scene and our calibrated energy-based model, whose energy level aligns with the success rate. 
The next best view is selected by maximizing the entropy reduction given the candidate view without any other heuristics. 
Our experiments, both in simulation and the real world, show that the proposed ActiveGrasp framework achieves the highest success rate given a limited view budget. We summarize our contributions as follows:

\begin{itemize}
    \item We give a clear definition of entropy of grasp distribution and derive the information gain of candidate views for the grasping task from an information-theoretic perspective.
    \item We estimate the entropy using the calibrated energy-based model and GAP such that the information gain estimation is accurate. 
    \item We present an active grasping benchmark with YCB~\cite{ycb} objects and a physically informed simulator, enabling reproducible results for further research in the community. 
\end{itemize}

\section{Related Works}
\label{sec:grasp-related}

\paragraph{Grasp Learning on Manifolds}
Robotic Grasping is a vibrant research problem and has been studied for a long time. 
We refer the readers to literature reviews for broader interests on the robotic grasping~\cite{kleeberger2020survey, zhang2022robotic, platt2022grasplearningmodelsmethods}.
Here, we mainly focus on grasp learning works on the SE(2) or SE(3) manifolds.
To learn grasp poses on the SE(2) manifolds, early works used Sample and Test methods~\cite{Pinto2016Supersizingselfsupervision, mahler2017dexnet20deeplearning, Lenz2015deeplearningfordetectinroboticgrasps} to uniformly sample candidates on the manifold and train a classifier to test each sample. 
Inspired by object detection in the computer vision community, later works like~\cite{Johns2016deeplearningagraspfuntionforgrasping, Chu2018realworldmultiobjectmultigraspdetection, Redmon2015realtimegraspdetectionusingconvolutionn} propose candidate regions and regress the parameters of the grasp pose on the manifold.
Learning grasp poses on the SE(3) manifold differs from learning on the SE(2) manifold in terms of network architecture, such as 3D convolution~\cite{vgn,Cai2022realtimecollisionfreegraspposedetection} and point models~\cite{Wei2021gpr}
To acquire a distribution of grasp poses, generative modeling of grasps on the manifold became popular recently.
Weng~\etal~\cite{Weng2024CAPGrasp} develop CAPGrasp, an $R^3 \times SO(2)$-equivariant grasp pose generative model under the assumption of approach-constrained grasp. 
For the SE(3) manifold, Lim~\cite{byeongdo2024equigraspflowse3equivariant6dofgrasppose} learns an SE(3) equivariant flow field to generate grasp poses on the SE(3) manifold.
Hu~\etal~\cite{Hu2024OrbitGrasp} maps each point in the cloud to a continuous grasp quality function over the 2-sphere $S^2$ and generates samples along the perpendicular direction of the surface normal.
The closest work to ours is Urain~\cite{se3diff}, which trains an energy-based model on the SE(3) manifold and runs the denoising steps to generate successful grasps. 
We leverage the energy function learned from~\cite{se3diff} to compute the information gain for next-best-view selection.

\paragraph{Active Perception}
Active Perception~\cite{bajcsy1988active,bajcsy2018revisiting} has long been a significant area of research in robotics and computer vision.
Various methods have been applied for uncertainty quantification for radiance field and 3DGS, including training neural networks~\cite{jin2023neu, upen,Ran2023neurar,guedon2022scone, guedon2023macarons, pan2022activenerf}, reinforcement learning~\cite{chaplot2020learning, ramakrishnan2020occupancy}, point cloud completion~\cite{dhami2023pred}.
Second-order information, such as Fisher Information, is also employed for uncertainty estimation and next-best-view selection~\cite{FisherRF,goli2023, jiang2025multimodal,yan2023active-neural-mapping}.

Active perception on robotic grasping has mostly focused on other data representations.
Gualtieri~\etal~\cite{marcusviewpointselectionforgraspdetection} utilized viewpoint direction for view selection.
Arruda~\etal~\cite{arrudaactivevisionfordexterousgraspingfornovelobjects} select views that could refine the surface reconstruction quality near the model-predicted contact points.
Breyer~\etal~\cite{breyer2022closedloop} selects the next best view by counting the number of previously occluded voxels in the candidate view.
Gregory~\etal~\cite{gregoryactiveexplorationusingtrajectoryoptimization} parametrize the occluded region as a mixture of Gaussians and minimize uncertainty within the region.
Chen~\etal~\cite{xiangyutransferableactivegraspingandrealembodieddataset} uses visibility as one of the reward signals to train an RL policy to select viewpoints.
Douglas~\etal~\cite{douglasmultiviewpicking} parametrize the model prediction as a Bernoulli distribution and compute entropy within 3D grids. They select views that observe the largest entropy grid cells.
Zhang~\etal~\cite{zhang2023ace} builds a tri-plane feature and trains a network to predict the affordance score in unobserved views and perform view selection among them.
Ma~\etal~\cite{ma2024activeNGF} uses the inconsistency between the prediction from the graspness and the TSDF feature planes as information gain for view selection.

\begin{figure*}[ht]
    \centering
    \includegraphics[width=0.9\linewidth]{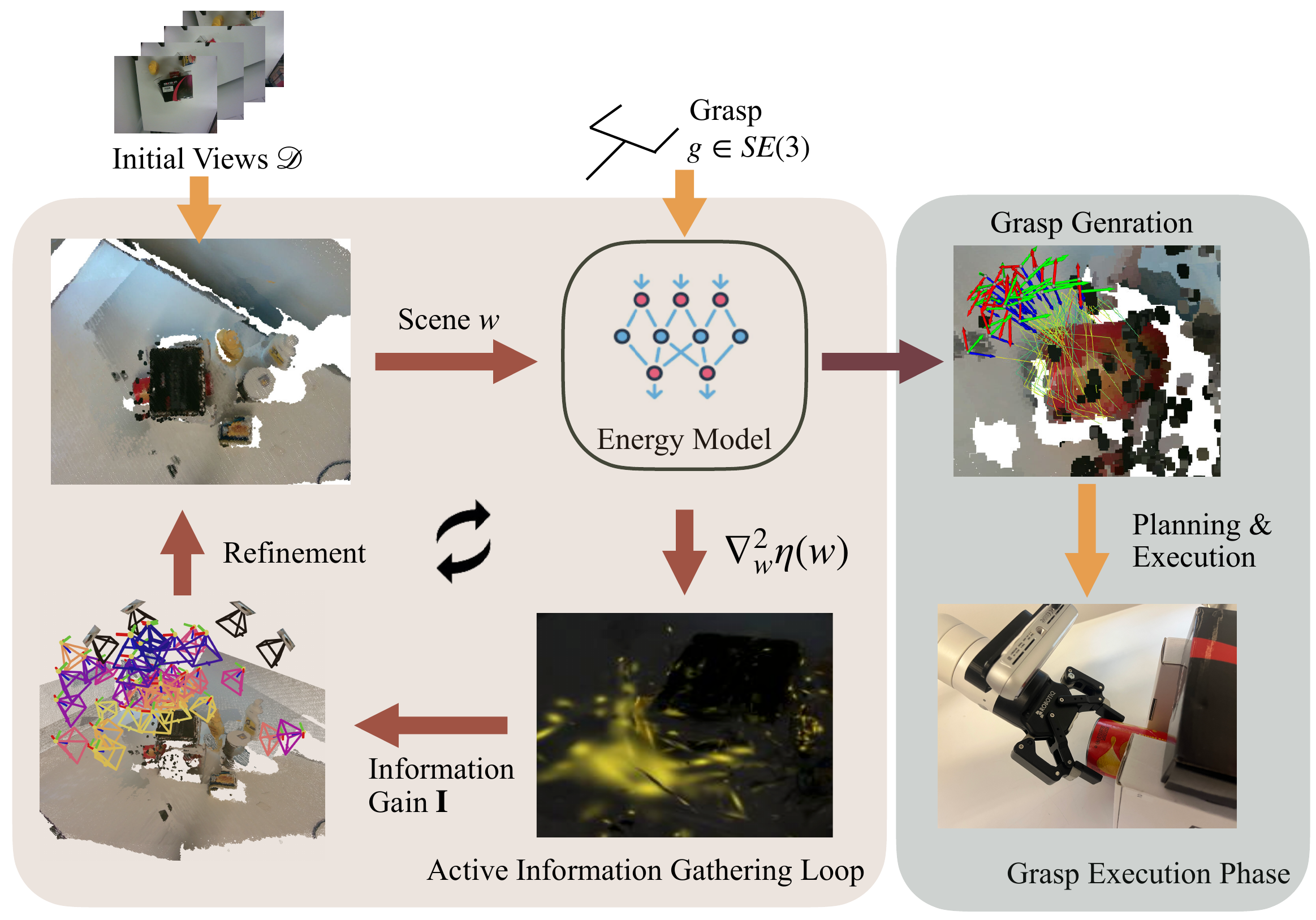}
    \caption{\textbf{Our Active Grasping Pipeline} Our method reconstructs 3D scene $w$ from a set of initial views. The energy-based model estimates the grasp entropy $\eta(w)$ using $w$ and the sampled grasp poses $g$. Then we compute the information gain $\mathbf{I}$ of candidate views from $\nabla_w^2\eta(w)$. We select the next best view with the highest information gain and repeat the procedure above until the view budget is met. Afterwards, we generate grasp poses using the refined scene representation. A planner generates paths to the grasps and executes them on a robot arm. }
    \label{fig:pipeline}
    \vspace{-5mm}
\end{figure*}

\paragraph{Model Calibration}
Model calibration aims to adjust a model so that its predicted probabilities better reflect the true likelihood of outcomes. 
Model calibration methods can be categorized into regularization, uncertainty estimation, and post-processing~\cite{jakob2023asurveyofuncertaintyindeepneuralnetwork}.
Regularization methods modify the training phase by label smoothing~\cite{christian2018rethinkingtheinception} or adding extra losses like weighted confidence-integrated loss~\cite{seonguk2019learningforsingleshotconfidencecalibration} or entropy term~\cite{gabriel2017regularizingneuralnetwork}.
Uncertainty estimation predicts an accurate estimation of the data uncertainty for calibration by using the Bayesian Network~\cite{Pavel2019subspaceinferenceforbayesiandeeplearning} or model ensembles~\cite{alireza202confidencecalibrationandpredictiveuncertainty}.
Post-processing tunes the model on an extra calibration set to align the prediction by temperature scaling~\cite{chugao2017oncalibrationofmodernneuralnetworks}, standard deviation scaling~\cite{dan2022evaluatingandcalibratinguncertainty}, or a Gaussian Process~\cite{Jonathan2020nonparametericcalibrationforclassification}.
One work close to ours is \cite{chugao2017oncalibrationofmodernneuralnetworks}, which tunes the temperature divided by logits on the calibration set. 
Kull~\etal~\cite{DirichletCalib} extend \cite{chugao2017oncalibrationofmodernneuralnetworks} using the Dirichlet distribution to map uncalibrated scores to calibrated scores.
Recently, Grathwohl~\etal~\cite{JEM} find that a classifier trained using the energy-based model method can automatically calibrate itself. 
However, they do not explore how to calibrate the energy-based model on the regression task, like grasp pose detection.
To the best of our knowledge, we are the first to calibrate the model on the grasp pose detection problem where the model's predicted success probability matches the real success rate.  


\label{sec:related}

\section{Method}
\label{sec:method}

\paragraph{Overview} Given a set of initial fixed views $\mathcal{D}$ and a target mask, ActiveGrasp first estimates the 3D scene representation $w$ and then computes the information gain $\mathbf{I}$ for $K$ candidate views using the energy-based model. We select the candidate view with the highest information gain, guide the robot to take one new observation at this view, and refine the scene $w$. This active information-gathering loop continues until the predetermined view budget is reached. Then we use the energy-based model to predict the grasp poses for the target. The whole pipeline is shown in Fig.~\ref{fig:pipeline}. We use 3D Gaussian Splatting~\cite{kerbl3dgaussians} (3DGS) as the scene representation. Each 3D Gaussian is parameterized with position, rotation, scale, RGB color, and a semantic channel indicating the target object. We introduce the Gaussian Approximation and the energy-based model in Sec~\ref{sec:preliminary}. We derive the information gain of candidate views with respect to the grasping task in Sec~\ref{sec:info-gain-grasp}. We show how to calibrate the energy level to compute the information gain in Sec~\ref{sec:EBM-calib}.

\subsection{Preliminary}
\label{sec:preliminary}
\paragraph{Gaussian Approximation of Posterior}
For an estimate of scene representation $\mathbf{W}$ from the training set $\mathcal{D}$, we can use a Gaussian distribution to approximate the distribution near the Maximum A Posteriori (MAP) estimate $w^*$ as \cite{kirsch2022unifying}
\begin{equation}
    p(w|\mathcal{D}) \sim \mathcal{N}(w^*,  \mathbf{H}''[w|\mathcal{D}]^{-1})
\label{eq:gaussian-approx}
\end{equation}
where $\mathbf{H}''(w|\mathcal{D})$ is the Hessian matrix for the Shannon entropy of the Posterior distribution $p(w|\mathcal{D})$. Using Bayes' Theorem and the additive property of logarithms, the Hessian of the posterior can be computed as 
\begin{equation}
     \mathbf{H}''[w|\mathcal{D}] = \mathbf{H}''[\mathcal{D}|w] + \mathbf{H}''[w]
     \label{eq:Hessian-compute}
\end{equation}
The Hessian can be approximated by the Jacobian using Gauss-Newton and estimated approximately diagonal due to the sparsity of the 3DGS representation~\cite{FisherRF}.
\begin{equation}
     \mathbf{H}''[w|\mathcal{D}] \approx \sum_{x\in\mathcal{D}}\text{diag}(\nabla_wf(x, w) \nabla_w f(x, w)^T)+ \lambda I    
\end{equation}
where $f(x,w)$ is the rendering equation at the camera pose $x$ and $\lambda$ is a regularizing constant. 

\paragraph{Energy-based Model}
An energy-based model~\cite{EBM} for grasping is a neural network that maps a pair of scene representation $w$ and grasp pose $g$ to a scalar: $E_\theta: w\times g \rightarrow \mathbb{R}$, where $w$ is the 3DGS model and $g \in SE(3)$. It models the conditional distribution of grasp poses given the scene  as:
\begin{equation}
    p(g  | w) = \frac{e^{- E_\theta(g , w)}}{Z} ; \quad Z = \int e^{-E_\theta(g , w)} dg
    \label{eq:ebm-definition}
\end{equation}
SE(3) Diffusion Field~\cite{se3diff} extends the model to condition on noise scale $\sigma_k$ as $E_\theta(g,\sigma_k, w)$ and trains the model on the SE(3) manifold using denoised score matching~\cite{ScoreMatching}. Noise is added to the ground-truth successful grasp poses as:
\begin{equation}
    \hat{g} = g \exp ( \sigma_k\hat{\epsilon}) \quad  \epsilon \sim \mathcal{N}(0, I)
\end{equation}
The model is trained with a combination of SDF Loss $\mathcal{L}_{\text{sdf}}$ and Denoised Score Matching Loss $\mathcal{L}_{\text{dsm}}$. The SDF Loss is the L1-Loss between the predicted Signed Distance Function (SDF) value on the reference points and the ground-truth SDF value on the same points. The ground truth SDF value is computed from depth images. The Denoised Score Matching Loss is:
\begin{equation}
\begin{aligned}
    &\mathcal{L}_{\text{dsm}} = \\ &\frac{1}{L} \sum_{k=0}^{L} \mathbb{E}_{g, \hat{g}} 
\left\| \left. \frac{\partial E_\theta (\gamma, \sigma_k, w)}{\partial \gamma} \right |_{\gamma = \hat{g}} 
+ 
\left. \frac{\partial \log \zeta(\gamma| g, \sigma_k \mathbf{I})}{\partial \gamma} \right |_{\gamma=\hat{g}} \right\|^2 
\end{aligned}
\label{eq:loss_sm}
\end{equation}
where the perturbed distribution $\zeta$ is:
\begin{equation}
    \zeta(\gamma| g, \sigma_k I) \propto \exp \left ( -\frac{1}{2} \; \|\text{Log} (g^{-1} \gamma) \|^2 / \sigma_k^2 \right)
\end{equation}
Annealed Langevin Dynamics~\cite{ScoreMatching} is used to sample on the $SE(3)$ manifold~\cite{se3diff}:
\begin{equation}
    g_{k-1} = g_k \exp  \left (-\alpha_k^2 \left. \frac{\partial E_\theta (\gamma, \sigma_k, w)}{\partial \gamma} \right |_{\gamma = g_k} + \alpha_k \epsilon \right )
\end{equation}
where $\alpha_k$ is a step-dependent coefficient.

\subsection{Expected Information Gain for Grasping}
\label{sec:info-gain-grasp}
We define the entropy of grasps as the conditional entropy $\mathbf{H}[S|G, W]$ where $G$, $W$, and $S$ are the random variables for the grasp pose, scene representation, and the final success event.
We denote this as a function of $w$ as $\eta(w)$:
\begin{equation}
    \eta(w) = \mathbf{H}[S|G, W] = \mathbb{E}_{p(g|w)}[h(g,w)]
    \label{eq:grasp-entropy-define}
\end{equation}
where $h(g, w)$ is the entropy of a single grasp pose $g$ for the given scene $w$. 
The success of a single grasp pose follows a Bernoulli distribution parameterized by $s(g,w)$. 
The entropy for this single grasp is defined as 
\begin{equation}
\begin{aligned}
    h(g,w) & = \mathrm{h}(s(g,w)) \\ 
           &= -s(g, w) \log s(g, w) - (1 - s(g, w)) \log (1-s(g, w))
\end{aligned}
\end{equation}
\begin{figure}[t]
    \centering
    \includegraphics[width=0.48\textwidth]{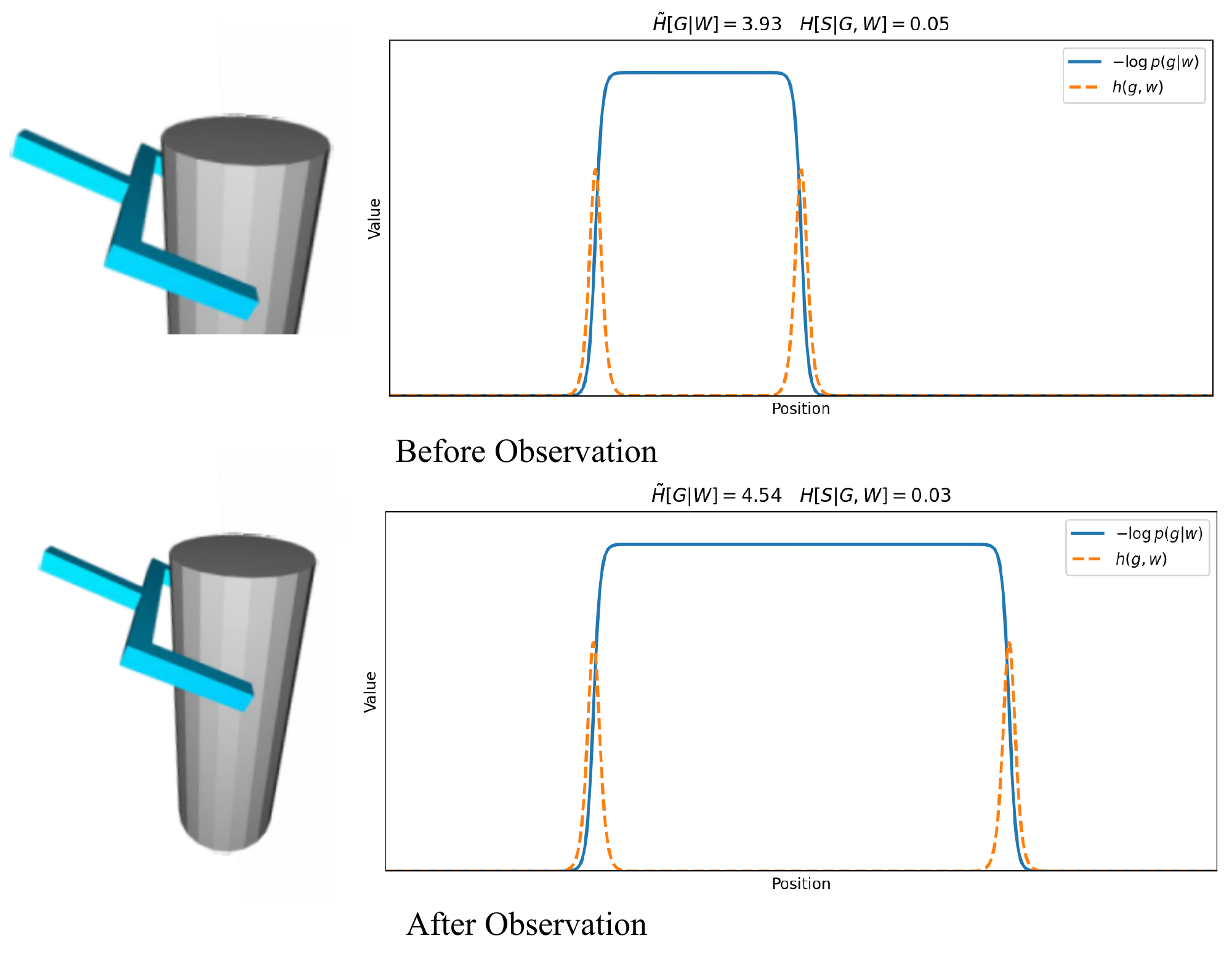}
    \caption{\textbf{Illustration of entropy on grasp poses} This is an example of the different definitions of entropy before the next observation (up) and after the next observation (down). We plot the success rate (blue) and entropy (dashed yellow) of a single grasp moving along the cup. $\tilde{\mathbf{H}}$ is the Shannon entropy of the grasp distribution and $\mathbf{H}$ is the conditional grasp entropy defined in our paper.}
    \label{fig:entropy-definition}
    \vspace{-5mm}
\end{figure}
We differentiate Eq.~\ref{eq:grasp-entropy-define} from the Shannon entropy of the grasp pose distribution 
\begin{equation}
    \tilde{\mathbf{H}}[G|W] = \mathbb{E}_{p(g|w)} [ - \log p(g|w) ]
\end{equation}
As is shown in the Fig.~\ref{fig:entropy-definition}, $\tilde{\mathbf{H}}$ increases after observation, which differs from the intuition that entropy should decrease after observation. On the other hand, $\mathbf{H}$ decreases due to more successful grasps being discovered.
When the scene representation is estimated from the observations, the entropy of the grasp pose is 
\begin{equation}
\begin{aligned}
    \mathbf{H}[S|G, W, \mathcal{D}] = \mathbb{E}_{p(w|\mathcal{D})} [ \eta(w) ]
\end{aligned}
\end{equation}
where $\eta(w) = \int  h(s, g) p(g|w) dg $. We apply Taylor expansion on the function $\eta(w)$ and only keep up to the second-order term. 
\begin{equation}
\begin{aligned}
    \eta(w) = \eta(w^*) &+ \nabla\eta^T(w^*)(w - w^*) \\ &+ \frac{1}{2} (w - w^*)^T \nabla_w^2
\eta(w^*)(w - w^*).
\end{aligned}
\label{eq:taylor}
\end{equation}
Using the Gaussian Approximation of the Posterior estimate from Eq.~\ref{eq:gaussian-approx}, the entropy can be approximated as: 
\begin{equation}
    \mathbf{H}[S|G, W, D]  = \eta(w^*) + \frac{1}{2}  \text{tr}  \left ( \nabla_w^2  \eta(w^*) \mathbf{H}''[w^*|\mathcal{D}]^{-1} \right ).
\label{eq:grasp-entropy}
\end{equation}
The information gain for the candidate views can be approximated as 
\begin{equation}
\begin{split}
    &\mathbf{I} [G,W; y_{\text{acq}} | x_{\text{acq}}, \mathcal{D}] \\
        &\approx   \frac{1}{2}  \text{tr}  \left \{ \nabla_w^2  \eta(w^*) \left [  \mathbf{H}''[w^*|\mathcal{D}]^{-1} -  \mathbf{H}''[w^*|x_{\text{acq}}, \mathcal{D}]^{-1} \right ] \right \}
\end{split} 
\label{eq:mutual-information}
\end{equation}
where $x_{\text{acq}}$ is the candidate view pose and $y_{\text{acq}}$ is the image at $x_{\text{acq}}$. We leave the proof of Eq.~\ref{eq:grasp-entropy} and Eq.~\ref{eq:mutual-information} to the appendix. The challenge to compute $\nabla_w^2 \eta(w)$ is the lack of a relationship between the success rate $s(g,w)$ and the density $p(g|w)$. Even though EBM estimates the density $p(g|w)$, it does not provide more information about the success rate. Therefore, we need to calibrate the energy level to the success rate, and we introduce this in Sec~\ref {sec:EBM-calib}. 

To compute $\nabla_w^2 \eta(w)$, we find that using the following approximation works well in practice.
\begin{equation}
    \nabla_w^2 \eta(w) = \nabla_w \eta(w) \nabla_w \eta(w)^T + \lambda I
\end{equation}
where $\nabla_w \eta(w)$ can be computed as:
\begin{equation}
\begin{aligned}
    \nabla_w \eta(w) = \mathbb{E}[\nabla_w  h] &- \mathbb{E}[h \nabla_wE_\theta(g, w, \sigma_k)] \\ &- \mathbb{E}[h] \mathbb{E}[ \nabla_wE_\theta(g, w, \sigma_k)] 
\end{aligned}
\label{eq:eta-first-order}
\end{equation}
The proof for Eq.~\ref{eq:eta-first-order} is left to the appendix.


\begin{table*}[ht]
\centering
\footnotesize
\resizebox{\linewidth}{!}{
\begin{tabular}{lcccccc}
\toprule
    Grasping Method + Active Learning Method & Success & Drop & Fail & Invalid & SR & ECE \\ \hline
ACE + ACE~\cite{zhang2023ace}    & 90 & 122 & 168 & 20 & 22.50\% & 0.35 \\ 
Contact Graspnet~\cite{sundermeyer2021contact} + ActiveNGF~\cite{ma2024activeNGF}   &   130 & 34  & 28 & 208 & 32.50\% & 0.32   \\
GSNet~\cite{GraspNet} + ActiveNGF~\cite{ma2024activeNGF}  & 249  & 59  & 36  & 62    & 62.00\%   & 0.30   \\
Se3diff~\cite{se3diff} + ActiveNGF~\cite{ma2024activeNGF} & 233 & 26 & 72& 69 & 58.25\% & 0.40 \\
Se3diff Scene + ActiveGrasp & 294 & 50 & 23 & 33 & 73.00\% & 0.28   \\  
Se3diff Calib + Random & 297 & 25 & 33 & 45 & 74.25\% & 0.06 \\
Se3diff Calib + Breyer~\cite{breyer2022closedloop} &  295 & 26& 27& 52 & 73.75\% & 0.05     \\
Se3diff Calib + FisherRF~\cite{FisherRF} & 294 & 35 & 26 & 45 & 73.50\%  &  0.05  \\
Se3diff Calib + ActiveNGF~\cite{ma2024activeNGF}   & 296     & 28   & 33   & 43      &  74.00\%   & 0.06       \\
Se3diff Calib + ACE~\cite{zhang2023ace}           &  296    & 27   & 31   & 45      &  74.00\%   & 0.07   \\
Se3diff Calib + $\text{Random}^\dagger$       &  306    & 29   & 29   & 38      &  76.50\%   & 0.05   \\
Ours                                               & 316     & 19   & 24   & 41      &  79.00\% &  0.02     \\ 
\bottomrule
\end{tabular}
}
\caption{ \textbf{Evaluation results for active grasping in simulated environments.} The methods are listed as grasping pose generation methods plus active learning methods. \textit{Se3diff Scene} is our scene augmented energy-based model without calibration. \textit{Se3diff Calib} is our calibrated energy-based model. \textit{ActiveGrasp} only selects the views with our proposed information gain. \textit{$\text{Random}^\dagger$} takes \textbf{8 more views} than all other active methods. \textbf{Drop} is defined as the robot finding the target but failing to lift it. \textbf{Fail} is defined as the robot fails to get in contact with the target, but the grasp model predicts feasible grasps. \textbf{Invalid} is defined as the grasp model that cannot predict any feasible grasps. \textbf{SR} is the success rate. $\textbf{ECE}$ is the Expected Calibration Error computed on all the executed grasps. }
\label{tab:sim-results}
\vspace{-5mm}
\end{table*}

\subsection{Calibrated Grasp Generation}
\label{sec:EBM-calib}
Prior work (JEM~\cite{JEM}) shows that EBMs can be calibrated for classification by combining Contrastive Divergence (CD) loss with classification losses, achieving a correspondence between energy and probability. 
However, applying CD loss directly to our setting is impractical because the conditioning variable $w$ is continuous and high-dimensional, making the CD computation prohibitively expensive. 
We therefore develop a practical calibration method specifically designed for grasp generation.

We formulate grasp success/failure as a two-category classification problem and train the network to output logits $(a_S, a_F)$ that parameterize success and failure probabilities:
\begin{equation}
    (a_S, a_F) = f_\theta(g, w, \sigma_k)
    \label{eq:network}
\end{equation}
The probabilities for the success and failure grasp poses are
\begin{equation}
    p_S = \frac{e^{a_S}}{e^{a_S} + e^{a_F} + 2}; \quad p_F = \frac{e^{a_F}}{e^{a_S} + e^{a_F} + 2}
\label{eq:dirich-prob}
\end{equation}
And the energy for success and failure is 
\begin{equation}
    E_S = - \log p_S ; \quad E_F = - \log p_F
\label{eq:energy}
\end{equation}
This formulation ensures that the energy is directly interpretable as a success probability, enabling reliable information gain computation.

A key insight is that the gradient of the energy function should point towards \emph{successful} grasps while also pointing \emph{away from} failure modes.
To achieve this, we employ \emph{dual score matching} losses: one that aligns the model's score with perturbations of successful grasps, and another that aligns with perturbations of failed grasps.
\begin{equation}
\begin{aligned}
    & \mathcal{L}^+ = \frac{1}{L} \sum_{k=0}^{L} \mathbb{E}_{g^+, \hat{g}} 
\left\|  \frac{\partial E_S (\mathbf{g}, w , \sigma_k)}{\partial \mathbf{g}}
+ 
\frac{\partial \log \zeta(\mathbf{g}| g^+, \sigma_k \mathbf{I})}{\partial \mathbf{g}} \right\|^2 \\ 
    & \mathcal{L}^- = \frac{1}{L} \sum_{k=0}^{L} \mathbb{E}_{g^-, \hat{g}} 
\left\|  \frac{\partial E_F (\mathbf{g}, w, \sigma_k)}{\partial \mathbf{g}}
+ 
\frac{\partial \log \zeta(\mathbf{g}| g^-, \sigma_k \mathbf{I})}{\partial \mathbf{g}} \right\|^2
\end{aligned}
\label{eq:loss_dual_sm}
\end{equation}
By learning from both successful and failed grasps, the model learns not only which grasps succeed, but also implicitly learns the failure modes, enabling the sampled grasps to avoid failure regions in pose space.

Unlike JEM~\cite{JEM}, simply adding a standard cross-entropy loss to enforce $p_S > p_F$ can be too restrictive, potentially disrupting the delicate energy structure learned by DSM.
Our key design choice is to use \emph{Average Precision (AP) Loss}~\cite{APLoss}, which provides only as much supervision as necessary: once positive samples consistently rank higher than negatives, the loss stops applying gradients.
This ``lazy'' supervisory signal preserves the score-matching structure while still enforcing calibration.
To compute AP Loss, we first create a set of equally spaced bins in [0,1] for the success and failure categories. 
For each grasp with its predicted probability $\{g_i, \boldsymbol{p}^i\}\;; \boldsymbol{p}=(p^i_S, p^i_F)$, we assign them to each of the bins with weight as:
\begin{equation}
    w^j_i = \max \left ( 0,  1 - \frac{|p^i - b_j|}{\Delta}\right)
\end{equation}
where $b_j$ is the center of j-th bin, $\Delta$ is the bin width, $p^i = p^i_S$ for success category and $p^i=p^i_F$ for failure category. 
Then we can compute the average precision for each category.
Since all the operations are differentiable, we can directly optimize the average precision for each category.

During training, we find that the energy level is unstable with both losses. We attribute the phenomenon to the fixed temperature, which is the default setting of energy-based models.
Comparing Eq.\ref{eq:ebm-definition} to a standard Boltzman definition, one finds that the EBM fixed the temperature as $T = 1$. 
By making the temperature learnable, the network can automatically adjust the energy scale to maintain numerical stability while still matching the score structure learned by DSM.
This is a simple but crucial modification that allows the calibration losses to converge reliably. 

In summary, the network is trained with the following loss:
\begin{equation}
    \mathcal{L}_{\mathrm{EBM}} = \frac{\mathcal{L}^+ + \mathcal{L}^-}{2} + \lambda_1 \mathcal{L}_{\text{ap}} + \lambda_2 \mathcal{L}_{\text{sdf}}
    \label{eq:loss}
\end{equation}

\section{Experiment}
\subsection{Implementation Details}
\paragraph{Energy-based Model Implementation}
The energy-based model takes scene $w$, grasp pose $g$, and noise level $\sigma_k$ as input. 
We first preprocess the Gaussian splats by taking only the means for each splat. We augment the energy-based model by fusing means and one-hot semantic vectors for each splat.
We use VNN~\cite{VectorNeurons} to extract features from the fused representation.
The fused representation for VNN is $\mathcal{P}\in\mathbb{R}^{N\times 2 \times 3}$ where $\mathcal{P}_i = \begin{bmatrix} p & l \end{bmatrix}^T \in \mathbb{R}^{2 \times 3}$. 
$p \in \mathbb{R}^3$ is the position of the point, and $l \in \mathbb{R}^3$ is all ones for foreground points and all zeros for background points.
We choose $N = 1024$.
The gripper is represented as a set of fixed anchor points in the gripper coordinate and then converted to the world coordinate by $g$. We use MLP to extract features from these anchor points. Then, features from scene and anchor points are concatenated in spatial dimension and sent to PointNet~\cite{PointNet}. The noise level $\sigma_k$ is embedded and fused with features in the PointNet. The output of the PointNet is sent to an MLP, which predicts $(a_S, a_F)$ in Eq.~\ref{eq:network}.  

\paragraph{Model Training}
We train the energy-based model on the Acronym Dataset~\cite{ACRONYM}. 
We randomly select object meshes and put them on the table in the simulation. 
Multiple views facing towards the target object are rendered on the sphere, centered on the target. 
We use Open3D~\cite{Open3D} to compute the ground truth SDF value of points near the target object surface. 
We choose $\lambda_1=1, \; \lambda_2 = 0.1$ for Eq.~\ref{eq:loss}. The model is trained for 200k steps with a batch size of 24. 
We use Adam optimizer with a learning rate of 1e-3. The noise schedule is chosen as $\sigma_k = (\sigma^{2k} -  1) / \log(\sigma), \; \sigma = 0.5, \; k \in [0, 1]$. 
The AP Loss is only computed for samples with noise level $\sigma_k$ when $k<0.01$. 

\paragraph{Information Gain Computation}
We first segment the foreground object from the background using 2D segmentation \cite{SAM2} for each capture image.
To initialize the 3DGS, we unproject the RGBD image to the 3D space and use it to initialize the 3DGS.
We use the forward kinematics of the robot arm to get the camera pose for each captured view to train the 3DGS. 
Each Gaussian is set to isotropic. 
The Spherical Harmonics degree is set to 0.
We train the 3DGS for 1k steps after we add one new view.
The 3DGS semantic channel is trained with cross-entropy loss on rendered segmentation maps and segmentation labels.
To filter out noise from 2D segmentation, we use DBSCAN~\cite{DBSCAN} to cluster all foreground splats with a minimum of 50 and a distance of 0.01. 
For the largest cluster of foreground splats, we take all the 3D Gaussians within a ball of radius $r$ at the center of the foreground 3D Gaussians.
Both the foreground and background 3D Gaussians are downsampled to 512 samples using farthest point sampling and concatenated as input for the energy-based model.
For candidate view proposal, we use Spherical Fibonacci Sampling~\cite{Benjamin2015sphericalfibonaccimapping} to sample poses on a sphere centering the foreground 3D Gaussians, with a radius uniformly sampled between 0.3 and 0.5 m.
The number of proposals is set to 128.

We compare our method against three different next-best-view planning methods in both simulation and the real world. For the grasp detection model, we compare against Contact GraspNet~\cite{sundermeyer2021contact}, ACE~\cite{zhang2023ace},  $SE(3)$ Diffusion Fields~\cite{se3diff} and GraspNet~\cite{GraspNet}. We use released checkpoints for Contact GraspNet and SE3Diffusion and train the ACE model from scratch using the official implementation. For the next-best-view selection, we also use official source code from Breyer~\cite{breyer2022closedloop}, 
 ACE~\cite{chen2024affordances} and ActiveNGF~\cite{ma2024activeNGF}

\begin{figure}[!t]
        \centering
        \includegraphics[width=\linewidth]{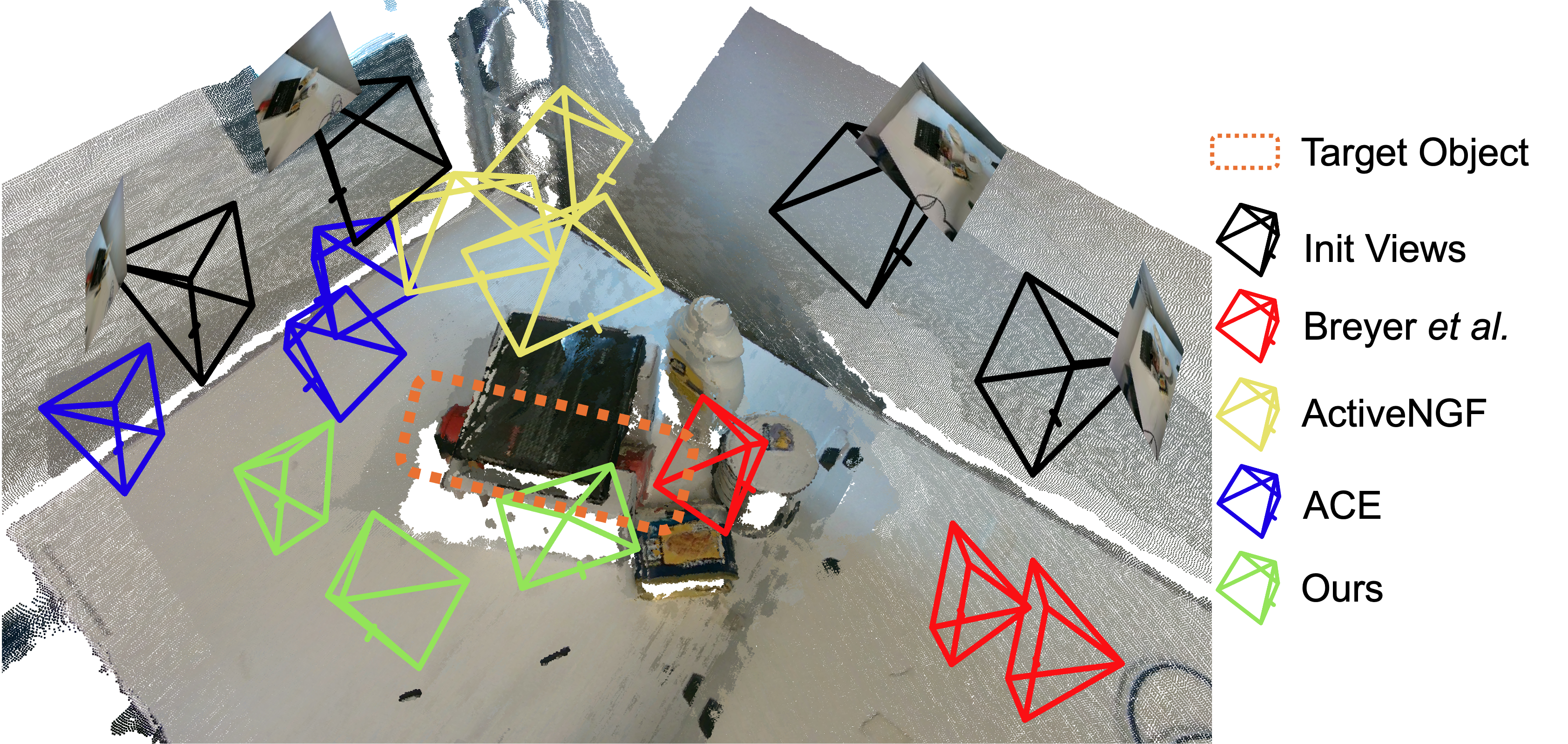}
        \caption{\textbf{Visualization for View Selections on Real-world Experiment} We showcase the top 3 view selections for different view selection methods at the same setup. Our method focuses more on perception for regions.}
        \label{fig:visualize}
        \vspace{-5mm}
\end{figure}

\subsection{Simulation Setup}
\label{sec:sim-setup}
We use the PyBullet simulation environment to evaluate our method in simulation \cite{pybullet}. To evaluate our method in simulation, we build a cluttered scene from the meshes of the objects in the YCB dataset~\cite{ycb}. 
A 7 DoF Franka robot is placed on one end of the table. 
We randomly generate configurations of these objects and place them into the scene. 
The number of objects in each scene is 10. 
One object is dropped inside the rectangular region in front of the robot by another. 
An RGBD camera is mounted on the wrist of the robot, with a resolution of 800x800 for both color and depth cameras. 
Examples of the simulation scene are shown in appendix.
We run all the experiments for 400 trials, including 20 different scenes, 5 different target objects in each scene, and 4 trials for each target. 
In each trial, the method takes 2 fixed views as initialization and actively selects 2 more views. 

\begin{table}[!t]
\centering
\footnotesize
\resizebox{0.95\linewidth}{!}{
    \begin{tabular}{m{2cm}|l|cccc}
\toprule
   Scene    &       Method       & Success & Drop & Fail & Invalid \\ \hline
 \multirow{3}{*}{\centering\includegraphics[width=0.8\linewidth]{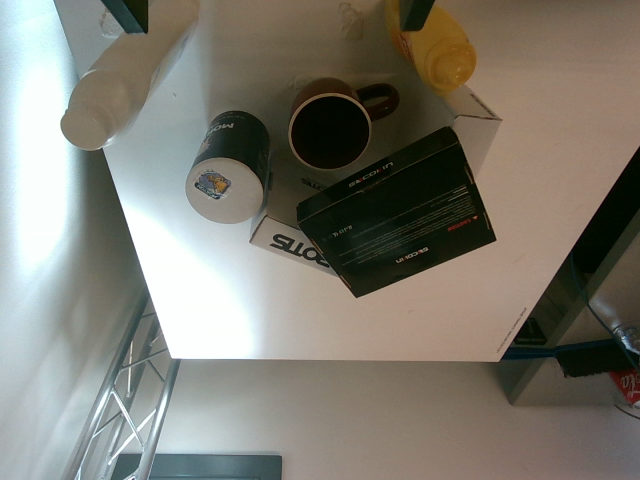} }  & Ours                   & 9/10      &  0/10   &  1/10  & 0/10   \\
                  & Breyer~\cite{breyer2022closedloop} & 6/10      & 2/10   & 2/10   & 0/10    \\
                  & ActiveNGF~\cite{ma2024activeNGF}      &  7/10      & 0/10    & 3/10   &  0/10    \\
                  & ACE~\cite{zhang2023ace}   &   8/10  &   0/10   &   2/10   &  0/10      \\ \hline
  \multirow{3}{*}{\centering\includegraphics[width=0.8\linewidth]{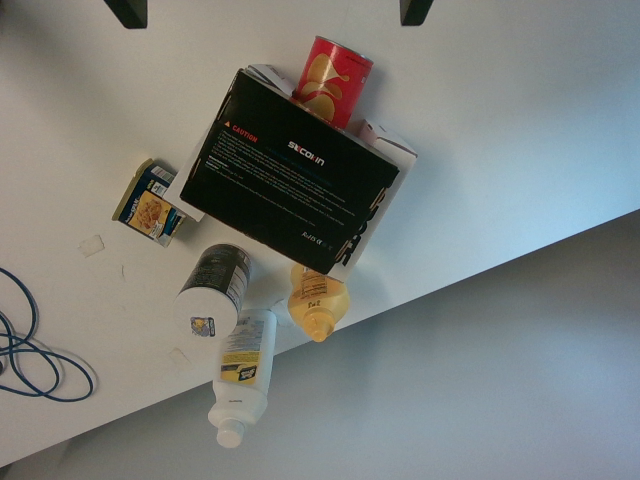} }  &    Ours  &    8/10      &    0/10     &   0/10   &   2/10     \\
                & Breyer~\cite{breyer2022closedloop} &  6/10      &  2/10  &  0/10  &  2/10   \\
                  & ActiveNGF~\cite{ma2024activeNGF}    &  6/10      &  0/10    & 0/10   & 4/10    \\
                  &  ACE~\cite{zhang2023ace}     &   4/10   &  2/10 &   2/10   &    2/10    \\ \hline
  \multirow{3}{*}{\centering\includegraphics[width=0.8\linewidth]{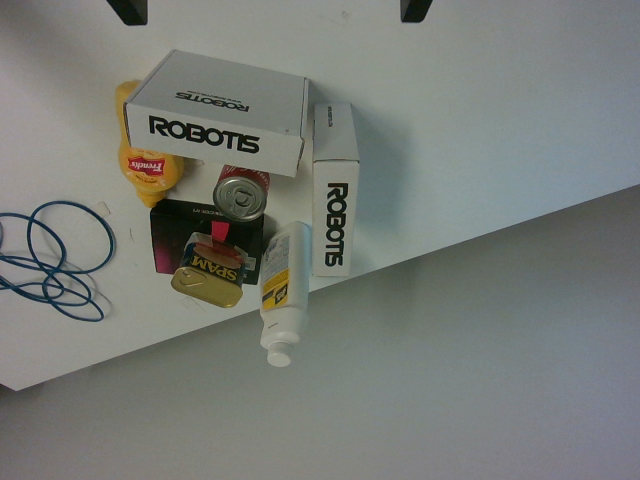} }  &     Ours                        &     9/10    &   0/10   &   1/10  &   0/10   \\
                & Breyer~\cite{breyer2022closedloop} & 8/10      & 2/10   &  0/10   &  0/10     \\
                  & ActiveNGF~\cite{ma2024activeNGF}       &  8/10      & 0/10    & 2/10   &  0/10     \\
                  &  ACE~\cite{zhang2023ace}     &   8/10     &  0/10    &   2/10   &  0/10       \\ 
\bottomrule
\end{tabular}
}

    \caption{\textbf{Evaluation results for active grasping in real world environments} The targets for three setups are red cup, red chip bottle, and red tomato can. }
    \label{tab:real}
    \vspace{-5mm}
\end{table}

\begin{table*}[!htbp]
\centering
\resizebox{0.98\textwidth}{!}{
\begin{tabular}{lccccccc}
\toprule
              & AP $\uparrow$  &  ECE $\downarrow$  & Ang (acc.) $\downarrow$  & Trans (acc.) $\downarrow$  & Ang (rec.) $\downarrow$  & Trans (rec.) $\downarrow$  & ECE (Bullet) $\downarrow$ \\ \hline
Se3diff~\cite{se3diff}  & 34.69 & 0.17 &   0.6244 & 0.0589 &  0.5233 & 0.0399 & 0.35    \\ 
Se3diff+Scene           & 66.00 & 0.15 & 0.4369       & 0.0546         & 0.3235       & 0.0184 & 0.30        \\ 
Se3diff+Scene+AP &  83.31 & 0.14 & 0.3517 & 0.0201 & 0.3138 & 0.0200 & 0.17    \\ 
Se3diff+Scene+FSM+AP &   83.13  & 0.03 & 0.4119 & 0.0238 & 0.3217 & 0.0189 & 0.08 \\ 
Se3diff+Scene+FSM+AP+T & 87.84  & 0.03 & 0.3764 & 0.0212 & 0.3068 & 0.0178 & 0.05    \\ 
\bottomrule
\end{tabular}
}
\caption{ \textbf{Quantitative comparison for grasping pose average precision on the ACRONYM dataset~\cite{ACRONYM}}. \textbf{AP}: average precision using the predicted energy. \textbf{ECE}: Expected Calibration Error on Acronym dataset. \textbf{Ang/Trans (acc.)}: Average closest angular and translational distance from the closest ground-truth successful grasps to the predicted grasps. \textbf{Ang/Trans(rec.)}: Average closest angular and translational distance from the predicted grasps to the closest ground-truth successful grasps. The unit of angular and translational distance is expressed as radians and meters. \textbf{ECE (Bullet)}: Expected Calibration Error after executing grasps in the Bullet simulator. \textbf{Scene}: Our implementation of model augmented with scene information. \textbf{FSM}: Model trained with both success and failure grasp. \textbf{AP} Model trained with AP Loss. \textbf{T}: Model trained with learnable temperature. }
\label{tab:acronym-ablation}
\vspace{-5mm}
\end{table*}

\subsection{Simulation Experiment}
\label{sec:sim-result}
We summarize the results of each grasping with the following metrics: \textbf{Success}: The number of successful trials. \textbf{Drop}: The number of trials when the robot is in touch with the target object but fails to firmly grasp it. \textbf{Fail}: The number of trials when there are feasible grasp poses, but the robot fails to be in contact with the target object. 
\textbf{Invalid}: The number of trials when the grasp prediction model cannot predict any grasp candidates or all the grasp candidates are rejected by the planner, due to infeasible kinematics or collision. The results are summarized in Tab~\ref{tab:sim-results}. 
Our calibrated energy-based model \textit{Se3diff-Calib} has a higher success rate than all previous grasp detection models~\cite {sundermeyer2021contact, GraspNet, chen2024affordances, se3diff}. 
We attribute this to the calibrated property of our energy-based model. Since the executed grasp is the grasp with the highest predicted success rate and the prediction is calibrated, it provides a guarantee for the success rate of execution. 
For next-based-view selection, we compare all previous methods using the same \textit{Se3diff-Calib}. 
Our method achieves the highest success rate of 79\%, higher than the best (74.25\%) of the other methods with same number of selected views.  
Meanwhile, we also find that our proposed view selection brings the best of the calibrated energy-based model, reaching the lowest calibration error of 0.02. 
We also compare our method with a setting using the same view selection algorithm but an uncalibrated energy-based model, as \textit{Se3diff Scene} + \textit{ActiveGrasp}, to show the effectiveness of energy calibration.
Finally, we compare with the setting, \textit{Se3diff Calib + $\text{Random}^\dagger$},  to show the effectiveness of next-best-view selection for grasping task. 
We let $\text{Random}^\dagger$ take \textbf{10 views} instead of 2 views. 
Even though taking more views could improve the success rate from 74.25\% to 76.50\%, it is still lower than our proposed active selection method with a higher execution time cost.


\subsection{Real World Experiment}
We run our real-world experiments on a Kinova 7 DoF Robot Arm. 
The robot is equipped with a parallel jaw end-effector gripper and a RealSense camera at the end joint. 
A set of objects from the YCB dataset will be placed in front of the robot. 
During the experiments, the robot takes 4 fixed views at the beginning and uses each method to actively select 2 views to build a representation of the scene.
Afterwards, the robot uses our calibrated energy-based model \textit{Se3diff Calib} to predict grasps and attempts to grasp the target object. 
We compare ActiveGrasp with a series of other grasp detection methods, including~\cite{vgn}, \cite{se3diff}, and~\cite{pas2017graspposedetectionpoint}. 
The results are summarized in Tab.~\ref{tab:real}. 
We visualize the results of the view selection in Fig.~\ref{fig:visualize} for the second setup. 
As is shown in the visualization, Breyer et al.~\cite{breyer2022closedloop} (red) selects views on the right-hand side of the object, where there are many obstacles. 
This is because they only rely on the coverage, failing to focus on the potential grasp location. ACE~\cite{chen2024affordances} (blue) and ActiveNGF \cite{ma2024activeNGF}~(yellow) can focus on the left-hand side, but their selections are close to the captured views (black) compared to our method~(green).

\subsection{Ablation Study}
\subsubsection{Calibrated Energy-based Model}
To show the effectiveness of our proposed calibrated energy-based model, we performed an ablation study on each module. 
We first evaluate our scene-aware grasp generation model \textit{Se3diff scene} and \textit{Se3diff} on the ACRONYM dataset~\cite{ACRONYM}. 
We use the validation set of the augmented grasp dataset for evaluation. 
We compute four metrics: \textbf{AP}, \textbf{ECE}, \textbf{Ang/Trans (acc.)} \textbf{Ang/Trans (rec.)}. 
We also randomly create scenes in the bullet simulator using YCB~\cite{ycb} objects as in Sec~\ref{sec:sim-setup} and place the gripper at the predicted grasp pose to grasp the object.
The result is regarded as the ground truth to compute \textbf{ECE (Bullet)}.
The result is summarized in Tab.~\ref{tab:acronym-ablation}. 
After adding scene information \textit{Scene} and AP Loss \textit{AP}, the model generates grasp poses closer to the ground truth distribution with lower distance measure \textbf{Ang/Trans (rec.)}, \textbf{Ang/Trans (acc.)}.
This is because it learns to distinguish good grasp from bad grasp as it has a higher average precision and lower calibration error on the dataset.
However, it still has high calibration error on its generation with large \textbf{ECE (Bullet)}.
After including the failure grasps in the score matching loss, the calibration for generation decreases, but the quality decreases since both \textbf{Ang/Trans (rec.)} and \textbf{Ang/Trans (acc.)} increase.
This issue is solved when we let the network learn the temperature, and the model performs better on all metrics.


\subsubsection{Run Time Analysis}
\label{sec:run-time}
To demonstrate the efficiency of our method in terms of runtime, we provide a breakdown of the time spent on different stages of our active grasping system. 
Our ActiveGrasp takes 9.61 ± 0.09 seconds for view selection, which is significantly faster than ACE~\cite{zhang2023ace} (18.24 ± 0.15 s), ActiveNGF~\cite{ma2024activeNGF} (27.66 ± 0.32 s), and Breyer~\etal~\cite{breyer2022closedloop} (32.97 ± 0.45 s).
Although vanilla FisherRF~\cite{FisherRF} (7.43 ± 0.01 s) is slightly faster than our method, it does not consider the grasping task and has a worse successful rate in our Experiments, as shown in Tab.~\ref{tab:sim-results}.

\section{Conclusion and Limitation}
\label{sec:conclusion}
In this paper, we addressed the challenging problem of active grasping, which improves robotic grasping quality by efficiently gathering information from the environment.
We proposed a novel, information-theoretic framework that leverages a calibrated energy-based model to compute information gain for the next-best-view (NBV) selection process in the grasping task.
We define the entropy of the grasp distribution and compute the entropy using the Gaussian Approximation of the Posterior of the scene and the energy-based model.
Our method eliminates the need for separate grasp affordance models or predefined heuristic metrics, offering a more principled and task-aware active perception system.
Our active grasping benchmark with physically informed simulators will also help follow-up work perform reproducible comparisons to boost the field's development further. 
The limitation of our work lies in the approximation of the second-order term $\nabla_w^2\eta(w)$ and the requirement of model calibration.
Future work could explore better approximation and calibration techniques given the domain shift.

{
    \small
{\small
    \paragraph{Acknowledgements}
    The authors gratefully appreciate support through the following grants: NSF FRR 2220868, NSF IIS-RI 2212433, and ONR N00014-22-1-2677.
}
    \bibliographystyle{ieeenat_fullname}
    \bibliography{main}

@String(CVPR= {IEEE Conf. Comput. Vis. Pattern Recog.})

@String(ICCV= {Int. Conf. Comput. Vis.})

@String(ECCV= {Eur. Conf. Comput. Vis.})

@String(ICLR = {Int. Conf. Learn. Represent.})

@String(AAAI = {AAAI})

@String(CVPR  = {CVPR})

@String(ICCV  = {ICCV})

@String(ECCV  = {ECCV})

@String(ICLR  = {ICLR})

@inproceedings{JEM,
  author       = {Will Grathwohl and
                  Kuan{-}Chieh Wang and
                  J{\"{o}}rn{-}Henrik Jacobsen and
                  David Duvenaud and
                  Mohammad Norouzi and
                  Kevin Swersky},
  title        = {Your classifier is secretly an energy based model and you should treat
                  it like one},
  booktitle    = {8th International Conference on Learning Representations, {ICLR} 2020,
                  Addis Ababa, Ethiopia, April 26-30, 2020},
  publisher    = {OpenReview.net},
  year         = {2020},
  url          = {https://openreview.net/forum?id=Hkxzx0NtDB},
  bibsource    = {dblp computer science bibliography, https://dblp.org}
}

@article{Benjamin2015sphericalfibonaccimapping,
  author       = {Benjamin Keinert and
                  Matthias Innmann and
                  Michael S{\"{a}}nger and
                  Marc Stamminger},
  title        = {Spherical fibonacci mapping},
  journal      = {{ACM} Trans. Graph.},
  volume       = {34},
  number       = {6},
  pages        = {193:1--193:7},
  year         = {2015},
  url          = {https://doi.org/10.1145/2816795.2818131},
  doi          = {10.1145/2816795.2818131},
  bibsource    = {dblp computer science bibliography, https://dblp.org}
}

@inproceedings{Wei2021gpr,
  author       = {Wei Wei and
                  Yongkang Luo and
                  Fuyu Li and
                  Guangyun Xu and
                  Jun Zhong and
                  Wanyi Li and
                  Peng Wang},
  title        = {{GPR:} Grasp Pose Refinement Network for Cluttered Scenes},
  booktitle    = {{IEEE} International Conference on Robotics and Automation, {ICRA}
                  2021, Xi'an, China, May 30 - June 5, 2021},
  pages        = {4295--4302},
  publisher    = {{IEEE}},
  year         = {2021},
  url          = {https://doi.org/10.1109/ICRA48506.2021.9561868},
  doi          = {10.1109/ICRA48506.2021.9561868},
  bibsource    = {dblp computer science bibliography, https://dblp.org}
}

@inproceedings{FisherRF,
  author       = {Wen Jiang and
                  Boshu Lei and
                  Kostas Daniilidis},
  editor       = {Ales Leonardis and
                  Elisa Ricci and
                  Stefan Roth and
                  Olga Russakovsky and
                  Torsten Sattler and
                  G{\"{u}}l Varol},
  title        = {FisherRF: Active View Selection and Mapping with Radiance Fields Using
                  Fisher Information},
  booktitle    = {Computer Vision - {ECCV} 2024 - 18th European Conference, Milan, Italy,
                  September 29-October 4, 2024, Proceedings, Part {XIII}},
  series       = {Lecture Notes in Computer Science},
  volume       = {15071},
  pages        = {422--440},
  publisher    = {Springer},
  year         = {2024},
  url          = {https://doi.org/10.1007/978-3-031-72624-8\_24},
  doi          = {10.1007/978-3-031-72624-8\_24},
  bibsource    = {dblp computer science bibliography, https://dblp.org}
}

@inproceedings{se3diff,
  author       = {Julen Urain and
                  Niklas Funk and
                  Jan Peters and
                  Georgia Chalvatzaki},
  title        = {SE(3)-DiffusionFields: Learning smooth cost functions for joint grasp
                  and motion optimization through diffusion},
  booktitle    = {{IEEE} International Conference on Robotics and Automation, {ICRA}
                  2023, London, UK, May 29 - June 2, 2023},
  pages        = {5923--5930},
  publisher    = {{IEEE}},
  year         = {2023},
  url          = {https://doi.org/10.1109/ICRA48891.2023.10161569},
  doi          = {10.1109/ICRA48891.2023.10161569},
  bibsource    = {dblp computer science bibliography, https://dblp.org}
}

@article{Weng2024CAPGrasp,
  author       = {Zehang Weng and
                  Haofei Lu and
                  Jens Lundell and
                  Danica Kragic},
  title        = {CAPGrasp: An {\textdollar}{\textbackslash}mathbb \{R\}{\^{}}\{3\}{\textbackslash}times
                  {\textbackslash}text\{SO(2)-Equivariant\}{\textdollar} Continuous
                  Approach-Constrained Generative Grasp Sampler},
  journal      = {{IEEE} Robotics Autom. Lett.},
  volume       = {9},
  number       = {4},
  pages        = {3641--3647},
  year         = {2024},
  url          = {https://doi.org/10.1109/lra.2024.3369444},
  doi          = {10.1109/LRA.2024.3369444},
  bibsource    = {dblp computer science bibliography, https://dblp.org}
}

@article{Lenz2015deeplearningfordetectinroboticgrasps,
  author       = {Ian Lenz and
                  Honglak Lee and
                  Ashutosh Saxena},
  title        = {Deep learning for detecting robotic grasps},
  journal      = {Int. J. Robotics Res.},
  volume       = {34},
  number       = {4-5},
  pages        = {705--724},
  year         = {2015},
  url          = {https://doi.org/10.1177/0278364914549607},
  doi          = {10.1177/0278364914549607},
  bibsource    = {dblp computer science bibliography, https://dblp.org}
}

@inproceedings{Redmon2015realtimegraspdetectionusingconvolutionn,
  author       = {Joseph Redmon and
                  Anelia Angelova},
  title        = {Real-time grasp detection using convolutional neural networks},
  booktitle    = {{IEEE} International Conference on Robotics and Automation, {ICRA}
                  2015, Seattle, WA, USA, 26-30 May, 2015},
  pages        = {1316--1322},
  publisher    = {{IEEE}},
  year         = {2015},
  url          = {https://doi.org/10.1109/ICRA.2015.7139361},
  doi          = {10.1109/ICRA.2015.7139361},
  bibsource    = {dblp computer science bibliography, https://dblp.org}
}

@article{Cai2022realtimecollisionfreegraspposedetection,
  author       = {Junhao Cai and
                  Jun Cen and
                  Haokun Wang and
                  Michael Yu Wang},
  title        = {Real-Time Collision-Free Grasp Pose Detection With Geometry-Aware
                  Refinement Using High-Resolution Volume},
  journal      = {{IEEE} Robotics Autom. Lett.},
  volume       = {7},
  number       = {2},
  pages        = {1888--1895},
  year         = {2022},
  url          = {https://doi.org/10.1109/LRA.2022.3142424},
  doi          = {10.1109/LRA.2022.3142424},
  bibsource    = {dblp computer science bibliography, https://dblp.org}
}

@inproceedings{Hu2024OrbitGrasp,
  author       = {Boce Hu and
                  Xupeng Zhu and
                  Dian Wang and
                  Zihao Dong and
                  Haojie Huang and
                  Chenghao Wang and
                  Robin Walters and
                  Robert Platt},
  editor       = {Pulkit Agrawal and
                  Oliver Kroemer and
                  Wolfram Burgard},
  title        = {OrbitGrasp: SE(3)-Equivariant Grasp Learning},
  booktitle    = {Conference on Robot Learning, 6-9 November 2024, Munich, Germany},
  series       = {Proceedings of Machine Learning Research},
  volume       = {270},
  pages        = {2456--2474},
  publisher    = {{PMLR}},
  year         = {2024},
  url          = {https://proceedings.mlr.press/v270/hu25b.html},
  bibsource    = {dblp computer science bibliography, https://dblp.org}
}

@inproceedings{Johns2016deeplearningagraspfuntionforgrasping,
  author       = {Edward Johns and
                  Stefan Leutenegger and
                  Andrew J. Davison},
  title        = {Deep learning a grasp function for grasping under gripper pose uncertainty},
  booktitle    = {2016 {IEEE/RSJ} International Conference on Intelligent Robots and
                  Systems, {IROS} 2016, Daejeon, South Korea, October 9-14, 2016},
  pages        = {4461--4468},
  publisher    = {{IEEE}},
  year         = {2016},
  url          = {https://doi.org/10.1109/IROS.2016.7759657},
  doi          = {10.1109/IROS.2016.7759657},
  bibsource    = {dblp computer science bibliography, https://dblp.org}
}

@article{Chu2018realworldmultiobjectmultigraspdetection,
  author       = {Fu{-}Jen Chu and
                  Ruinian Xu and
                  Patricio A. Vela},
  title        = {Real-World Multiobject, Multigrasp Detection},
  journal      = {{IEEE} Robotics Autom. Lett.},
  volume       = {3},
  number       = {4},
  pages        = {3355--3362},
  year         = {2018},
  url          = {https://doi.org/10.1109/LRA.2018.2852777},
  doi          = {10.1109/LRA.2018.2852777},
  bibsource    = {dblp computer science bibliography, https://dblp.org}
}

@InProceedings{	  arrudaactivevisionfordexterousgraspingfornovelobjects,
  author	= {Arruda, Ermano and Wyatt, Jeremy and Kopicki, Marek},
  booktitle	= {2016 IEEE/RSJ International Conference on Intelligent
		  Robots and Systems (IROS)},
  title		= {Active vision for dexterous grasping of novel objects},
  year		= {2016},
  volume	= {},
  number	= {},
  pages		= {2881-2888},
  doi		= {10.1109/IROS.2016.7759446}
}

@InProceedings{	  marcusviewpointselectionforgraspdetection,
  author	= {Marcus Gualtieri and Robert Platt Jr.},
  title		= {Viewpoint selection for grasp detection},
  booktitle	= {2017 {IEEE/RSJ} International Conference on Intelligent
		  Robots and Systems, {IROS} 2017, Vancouver, BC, Canada,
		  September 24-28, 2017},
  pages		= {258--264},
  publisher	= {{IEEE}},
  year		= {2017},
  url		= {https://doi.org/10.1109/IROS.2017.8202166},
  doi		= {10.1109/IROS.2017.8202166},
  bibsource	= {dblp computer science bibliography, https://dblp.org}
}

@article{EBM,
  title={A tutorial on energy-based learning},
  author={LeCun, Yann and Chopra, Sumit and Hadsell, Raia and Ranzato, M and Huang, Fujie and others},
  journal={Predicting structured data},
  volume={1},
  number={0},
  year={2006}
}

@article{levinelearningrobotgrasping2018,
  author       = {Sergey Levine and
                  Peter Pastor and
                  Alex Krizhevsky and
                  Julian Ibarz and
                  Deirdre Quillen},
  title        = {Learning hand-eye coordination for robotic grasping with deep learning
                  and large-scale data collection},
  journal      = {Int. J. Robotics Res.},
  volume       = {37},
  number       = {4-5},
  pages        = {421--436},
  year         = {2018},
  url          = {https://doi.org/10.1177/0278364917710318},
  doi          = {10.1177/0278364917710318},
  bibsource    = {dblp computer science bibliography, https://dblp.org}
}

@inproceedings{robertplatthighprecisiongraspindense2016,
  author       = {Marcus Gualtieri and
                  Andreas ten Pas and
                  Kate Saenko and
                  Robert Platt Jr.},
  title        = {High precision grasp pose detection in dense clutter},
  booktitle    = {2016 {IEEE/RSJ} International Conference on Intelligent Robots and
                  Systems, {IROS} 2016, Daejeon, South Korea, October 9-14, 2016},
  pages        = {598--605},
  publisher    = {{IEEE}},
  year         = {2016},
  url          = {https://doi.org/10.1109/IROS.2016.7759114},
  doi          = {10.1109/IROS.2016.7759114},
  bibsource    = {dblp computer science bibliography, https://dblp.org}
}

@STRING{aaai	= {AAAI} }

@STRING{cvpr	= {IEEE Conf. Comput. Vis. Pattern Recog.} }

@STRING{cvpr	= {CVPR} }

@STRING{eccv	= {Eur. Conf. Comput. Vis.} }

@STRING{eccv	= {ECCV} }

@STRING{iccv	= {Int. Conf. Comput. Vis.} }

@STRING{iccv	= {ICCV} }

@STRING{iclr	= {Int. Conf. Learn. Represent.} }

@STRING{iclr	= {ICLR} }

@inproceedings{PointNet,
  author       = {Charles Ruizhongtai Qi and
                  Hao Su and
                  Kaichun Mo and
                  Leonidas J. Guibas},
  title        = {PointNet: Deep Learning on Point Sets for 3D Classification and Segmentation},
  booktitle    = {2017 {IEEE} Conference on Computer Vision and Pattern Recognition,
                  {CVPR} 2017, Honolulu, HI, USA, July 21-26, 2017},
  pages        = {77--85},
  publisher    = {{IEEE} Computer Society},
  year         = {2017},
  url          = {https://doi.org/10.1109/CVPR.2017.16},
  doi          = {10.1109/CVPR.2017.16},
  bibsource    = {dblp computer science bibliography, https://dblp.org}
}

@Article{	  bajcsy1988active,
  title		= {Active perception},
  author	= {Bajcsy, Ruzena},
  journal	= {Proceedings of the IEEE},
  volume	= {76},
  number	= {8},
  pages		= {966--1005},
  year		= {1988},
  publisher	= {IEEE}
}

@Article{	  bajcsy2018revisiting,
  title		= {Revisiting active perception},
  author	= {Bajcsy, Ruzena and Aloimonos, Yiannis and Tsotsos, John
		  K},
  journal	= {Autonomous Robots},
  volume	= {42},
  pages		= {177--196},
  year		= {2018},
  publisher	= {Springer}
}

@InProceedings{	  breyer2022closedloop,
  author	= {Michel Breyer and Lionel Ott and Roland Siegwart and Jen
		  Jen Chung},
  title		= {Closed-Loop Next-Best-View Planning for Target-Driven
		  Grasping},
  booktitle	= {{IEEE/RSJ} International Conference on Intelligent Robots
		  and Systems, {IROS} 2022, Kyoto, Japan, October 23-27,
		  2022},
  pages		= {1411--1416},
  publisher	= {{IEEE}},
  year		= {2022},
  url		= {https://doi.org/10.1109/IROS47612.2022.9981472},
  doi		= {10.1109/IROS47612.2022.9981472},
  bibsource	= {dblp computer science bibliography, https://dblp.org}
}

@inproceedings{chaplot2020learning,
  title={Learning To Explore Using Active Neural SLAM},
  author={Chaplot, Devendra Singh and Gandhi, Dhiraj and Gupta, Saurabh and Gupta, Abhinav and Salakhutdinov, Ruslan},
  booktitle={International Conference on Learning Representations (ICLR)},
  year={2020}
}

@inproceedings{chen2024affordances,
  title={Affordances-Oriented Planning using Foundation Models for Continuous Vision-Language Navigation},
  author={Chen, Jiaqi and Lin, Bingqian and Liu, Xinmin and Ma, Lin and Liang, Xiaodan and Wong, Kwan-Yee~K.},
  booktitle = "Proceedings of the AAAI Conference on Artificial Intelligence",
  year={2025}
}

@inproceedings{SAM2,
  title={{SAM} 2: Segment Anything in Images and Videos},
  author={Nikhila Ravi and Valentin Gabeur and Yuan-Ting Hu and Ronghang Hu and Chaitanya Ryali and Tengyu Ma and Haitham Khedr and Roman R{\"a}dle and Chloe Rolland and Laura Gustafson and Eric Mintun and Junting Pan and Kalyan Vasudev Alwala and Nicolas Carion and Chao-Yuan Wu and Ross Girshick and Piotr Dollar and Christoph Feichtenhofer},
  booktitle={The Thirteenth International Conference on Learning Representations},
  year={2025},
  url={https://openreview.net/forum?id=Ha6RTeWMd0}
}

@inproceedings{VectorNeurons,
  author       = {Congyue Deng and
                  Or Litany and
                  Yueqi Duan and
                  Adrien Poulenard and
                  Andrea Tagliasacchi and
                  Leonidas J. Guibas},
  title        = {Vector Neurons: {A} General Framework for SO(3)-Equivariant Networks},
  booktitle    = {2021 {IEEE/CVF} International Conference on Computer Vision, {ICCV}
                  2021, Montreal, QC, Canada, October 10-17, 2021},
  pages        = {12180--12189},
  publisher    = {{IEEE}},
  year         = {2021},
  url          = {https://doi.org/10.1109/ICCV48922.2021.01198},
  doi          = {10.1109/ICCV48922.2021.01198},
  bibsource    = {dblp computer science bibliography, https://dblp.org}
}

@InProceedings{	  dhami2023pred,
  title		= {Pred-nbv: Prediction-guided next-best-view planning for 3d
		  object reconstruction},
  author	= {Dhami, Harnaik and Sharma, Vishnu D and Tokekar, Pratap},
  booktitle	= {2023 IEEE/RSJ International Conference on Intelligent
		  Robots and Systems (IROS)},
  pages		= {7149--7154},
  year		= {2023},
  organization	= {IEEE}
}

@InProceedings{	  douglasmultiviewpicking,
  author	= {Douglas Morrison and Peter Corke and J{\"{u}}rgen Leitner},
  title		= {Multi-View Picking: Next-best-view Reaching for Improved
		  Grasping in Clutter},
  booktitle	= {International Conference on Robotics and Automation,
		  {ICRA} 2019, Montreal, QC, Canada, May 20-24, 2019},
  pages		= {8762--8768},
  publisher	= {{IEEE}},
  year		= {2019},
  url		= {https://doi.org/10.1109/ICRA.2019.8793805},
  doi		= {10.1109/ICRA.2019.8793805},
  bibsource	= {dblp computer science bibliography, https://dblp.org}
}

@inproceedings{ACRONYM,
  author       = {Clemens Eppner and
                  Arsalan Mousavian and
                  Dieter Fox},
  title        = {{ACRONYM:} {A} Large-Scale Grasp Dataset Based on Simulation},
  booktitle    = {{IEEE} International Conference on Robotics and Automation, {ICRA}
                  2021, Xi'an, China, May 30 - June 5, 2021},
  pages        = {6222--6227},
  publisher    = {{IEEE}},
  year         = {2021},
  url          = {https://doi.org/10.1109/ICRA48506.2021.9560844},
  doi          = {10.1109/ICRA48506.2021.9560844},
  bibsource    = {dblp computer science bibliography, https://dblp.org}
}

@article{jakob2023asurveyofuncertaintyindeepneuralnetwork,
  author       = {Jakob Gawlikowski and
                  Cedrique Rovile Njieutcheu Tassi and
                  Mohsin Ali and
                  Jongseok Lee and
                  Matthias Humt and
                  Jianxiang Feng and
                  Anna M. Kruspe and
                  Rudolph Triebel and
                  Peter Jung and
                  Ribana Roscher and
                  Muhammad Shahzad and
                  Wen Yang and
                  Richard Bamler and
                  Xiaoxiang Zhu},
  title        = {A survey of uncertainty in deep neural networks},
  journal      = {Artif. Intell. Rev.},
  volume       = {56},
  number       = {{S1}},
  pages        = {1513--1589},
  year         = {2023},
  url          = {https://doi.org/10.1007/s10462-023-10562-9},
  doi          = {10.1007/S10462-023-10562-9},
  bibsource    = {dblp computer science bibliography, https://dblp.org}
}

@inproceedings{christian2018rethinkingtheinception,
  author       = {Christian Szegedy and
                  Vincent Vanhoucke and
                  Sergey Ioffe and
                  Jonathon Shlens and
                  Zbigniew Wojna},
  title        = {Rethinking the Inception Architecture for Computer Vision},
  booktitle    = {2016 {IEEE} Conference on Computer Vision and Pattern Recognition,
                  {CVPR} 2016, Las Vegas, NV, USA, June 27-30, 2016},
  pages        = {2818--2826},
  publisher    = {{IEEE} Computer Society},
  year         = {2016},
  url          = {https://doi.org/10.1109/CVPR.2016.308},
  doi          = {10.1109/CVPR.2016.308},
  bibsource    = {dblp computer science bibliography, https://dblp.org}
}

@Article{	  goli2023,
  title		= {{Bayes' Rays}: Uncertainty Quantification in Neural
		  Radiance Fields},
  author	= {Lily Goli and Cody Reading and Silvia Sellán and Alec
		  Jacobson and Andrea Tagliasacchi},
  journal	= {arXiv},
  year		= {2023}
}

@InProceedings{	  gregoryactiveexplorationusingtrajectoryoptimization,
  author	= {Gregory Kahn and Peter Sujan and Sachin Patil and Shaunak
		  D. Bopardikar and Julian Ryde and Kenneth Y. Goldberg and
		  Pieter Abbeel},
  title		= {Active exploration using trajectory optimization for
		  robotic grasping in the presence of occlusions},
  booktitle	= {{IEEE} International Conference on Robotics and
		  Automation, {ICRA} 2015, Seattle, WA, USA, 26-30 May,
		  2015},
  pages		= {4783--4790},
  publisher	= {{IEEE}},
  year		= {2015},
  url		= {https://doi.org/10.1109/ICRA.2015.7139864},
  doi		= {10.1109/ICRA.2015.7139864},
  bibsource	= {dblp computer science bibliography, https://dblp.org}
}

@Article{	  guedon2022scone,
  title		= {SCONE: Surface Coverage Optimization in Unknown
		  Environments by Volumetric Integration},
  author	= {Gu{\'e}don, Antoine and Monasse, Pascal and Lepetit,
		  Vincent},
  journal	= {Advances in Neural Information Processing Systems},
  volume	= {35},
  pages		= {20731--20743},
  year		= {2022}
}

@InProceedings{	  guedon2023macarons,
  title		= {MACARONS: Mapping And Coverage Anticipation with RGB
		  Online Self-Supervision},
  author	= {Gu{\'e}don, Antoine and Monnier, Tom and Monasse, Pascal
		  and Lepetit, Vincent},
  booktitle	= {Proceedings of the IEEE/CVF Conference on Computer Vision
		  and Pattern Recognition},
  pages		= {940--951},
  year		= {2023}
}

@inproceedings{jin2023neu,
  title={Neu-nbv: Next best view planning using uncertainty estimation in image-based neural rendering},
  author={Jin, Liren and Chen, Xieyuanli and R{\"u}ckin, Julius and Popovi{\'c}, Marija},
  booktitle={2023 IEEE/RSJ International Conference on Intelligent Robots and Systems (IROS)},
  pages={11305--11312},
  year={2023},
  organization={IEEE}
}

@Article{	  kerbl3dgaussians,
  author	= {Kerbl, Bernhard and Kopanas, Georgios and Leimkuhler,
		  Thomas and Drettakis, George},
  title		= {3D Gaussian Splatting for Real-Time Radiance Field
		  Rendering},
  journal	= {ACM Transactions on Graphics},
  number	= {4},
  volume	= {42},
  month		= {July},
  year		= {2023},
  url		= {https://repo-sam.inria.fr/fungraph/3d-gaussian-splatting/}
}

@Article{	  kirsch2022unifying,
  title		= {Unifying Approaches in Active Learning and Active Sampling
		  via Fisher Information and Information-Theoretic
		  Quantities},
  author	= {Andreas Kirsch and Yarin Gal},
  journal	= {Transactions on Machine Learning Research},
  issn		= {2835-8856},
  year		= {2022},
  url		= {https://openreview.net/forum?id=UVDAKQANOW},
  note		= {Expert Certification}
}

@InProceedings{	  ma2024activeNGF,
  title		= {Active Perception for Grasp Detection via Neural Graspness
		  Field},
  author	= {Haoxiang Ma and Modi Shi and Boyang Gao and Di Huang},
  booktitle	= {The Thirty-eighth Annual Conference on Neural Information
		  Processing Systems},
  year		= {2024},
  url		= {https://openreview.net/forum?id=6FYh6gxzPf}
}

@InProceedings{		  mahler2017dexnet20deeplearning,
  title		= {Dex-Net 2.0: Deep Learning to Plan Robust Grasps with
		  Synthetic Point Clouds and Analytic Grasp Metrics},
  author	= {Jeffrey Mahler and Jacky Liang and Sherdil Niyaz and
		  Michael Laskey and Richard Doan and Xinyu Liu and Juan
		  Aparicio Ojea and Ken Goldberg},
  year		= {2017},
  booktitle = {RSS}
}

@InProceedings{	  pan2022activenerf,
  title		= {ActiveNeRF: Learning Where to See with Uncertainty
		  Estimation},
  author	= {Pan, Xuran and Lai, Zihang and Song, Shiji and Huang,
		  Gao},
  booktitle	= {ECCV},
  pages		= {230--246},
  year		= {2022},
  organization	= {Springer}
}

@article{pas2017graspposedetectionpoint,
  title={Grasp pose detection in point clouds},
  author={Ten Pas, Andreas and Gualtieri, Marcus and Saenko, Kate and Platt, Robert},
  journal={The International Journal of Robotics Research},
  volume={36},
  number={13-14},
  pages={1455--1473},
  year={2017},
  publisher={SAGE Publications Sage UK: London, England}
}

@article{platt2022grasplearningmodelsmethods,
  title={Grasp learning: Models, methods, and performance},
  author={Platt, Robert},
  journal={Annual Review of Control, Robotics, and Autonomous Systems},
  volume={6},
  number={1},
  pages={363--389},
  year={2023},
  publisher={Annual Reviews}
}

@MISC{pybullet,
author =   {Erwin Coumans and Yunfei Bai},
title =    {PyBullet, a Python module for physics simulation for games, robotics and machine learning},
howpublished = {\url{http://pybullet.org}},
year = {2016--2021}
}

@InProceedings{	  ramakrishnan2020occupancy,
  title		= {Occupancy anticipation for efficient exploration and
		  navigation},
  author	= {Ramakrishnan, Santhosh K and Al-Halah, Ziad and Grauman,
		  Kristen},
  booktitle	= {Computer Vision--ECCV 2020: 16th European Conference,
		  Glasgow, UK, August 23--28, 2020, Proceedings, Part V 16},
  pages		= {400--418},
  year		= {2020},
  organization	= {Springer}
}

@Article{	  ran2023neurar,
  title		= {NeurAR: Neural Uncertainty for Autonomous 3D
		  Reconstruction With Implicit Neural Representations},
  volume	= {8},
  issn		= {2377-3774},
  url		= {http://dx.doi.org/10.1109/LRA.2023.3235686},
  doi		= {10.1109/lra.2023.3235686},
  number	= {2},
  journal	= {IEEE Robotics and Automation Letters},
  publisher	= {Institute of Electrical and Electronics Engineers (IEEE)},
  author	= {Ran, Yunlong and Zeng, Jing and He, Shibo and Chen, Jiming
		  and Li, Lincheng and Chen, Yingfeng and Lee, Gimhee and Ye,
		  Qi},
  year		= {2023},
  month		= feb,
  pages		= {1125–1132}
}

@inproceedings{ScoreMatching,
  author       = {Yang Song and
                  Stefano Ermon},
  editor       = {Hanna M. Wallach and
                  Hugo Larochelle and
                  Alina Beygelzimer and
                  Florence d'Alch{\'{e}}{-}Buc and
                  Emily B. Fox and
                  Roman Garnett},
  title        = {Generative Modeling by Estimating Gradients of the Data Distribution},
  booktitle    = {Advances in Neural Information Processing Systems 32: Annual Conference
                  on Neural Information Processing Systems 2019, NeurIPS 2019, December
                  8-14, 2019, Vancouver, BC, Canada},
  pages        = {11895--11907},
  year         = {2019},
  url          = {https://proceedings.neurips.cc/paper/2019/hash/3001ef257407d5a371a96dcd947c7d93-Abstract.html},
  bibsource    = {dblp computer science bibliography, https://dblp.org}
}

@InProceedings{	  upen,
  title		= {Uncertainty-driven planner for exploration and
		  navigation},
  author	= {Georgakis, Georgios and Bucher, Bernadette and Arapin,
		  Anton and Schmeckpeper, Karl and Matni, Nikolai and
		  Daniilidis, Kostas},
  booktitle	= {ICRA},
  year		= {2022}
}

@InProceedings{	  vgn,
  title		= {Volumetric grasping network: Real-time 6 dof grasp
		  detection in clutter},
  author	= {Breyer, Michel and Chung, Jen Jen and Ott, Lionel and
		  Siegwart, Roland and Nieto, Juan},
  booktitle	= {Conference on Robot Learning},
  pages		= {1602--1611},
  year		= {2021},
  organization	= {PMLR}
}

@InProceedings{	  xiangyutransferableactivegraspingandrealembodieddataset,
  author	= {Xiangyu Chen and Zelin Ye and Jiankai Sun and Yuda Fan and
		  Fang Hu and Chenxi Wang and Cewu Lu},
  title		= {Transferable Active Grasping and Real Embodied Dataset},
  booktitle	= {2020 {IEEE} International Conference on Robotics and
		  Automation, {ICRA} 2020, Paris, France, May 31 - August 31,
		  2020},
  pages		= {3611--3618},
  publisher	= {{IEEE}},
  year		= {2020},
  url		= {https://doi.org/10.1109/ICRA40945.2020.9197185},
  doi		= {10.1109/ICRA40945.2020.9197185},
  bibsource	= {dblp computer science bibliography, https://dblp.org}
}

@InProceedings{	  yan2023active-neural-mapping,
  title		= {Active Neural Mapping},
  author	= {Zike Yan and Haoxiang Yang and Hongbin Zha},
  year		= {2023},
  booktitle	= {ICCV}
}

@InProceedings{	  zhang2023ace,
  author	= {Xuechao Zhang and Dong Wang and Sun Han and Weichuang Li
		  and Bin Zhao and Zhigang Wang and Xiaoming Duan and
		  Chongrong Fang and Xuelong Li and Jianping He},
  editor	= {Jie Tan and Marc Toussaint and Kourosh Darvish},
  title		= {Affordance-Driven Next-Best-View Planning for Robotic
		  Grasping},
  booktitle	= {Conference on Robot Learning, CoRL 2023, 6-9 November
		  2023, Atlanta, GA, {USA}},
  series	= {Proceedings of Machine Learning Research},
  volume	= {229},
  pages		= {2849--2862},
  publisher	= {{PMLR}},
  year		= {2023},
  url		= {https://proceedings.mlr.press/v229/zhang23i.html},
  bibsource	= {dblp computer science bibliography, https://dblp.org}
}

@inproceedings{sundermeyer2021contact,
  title={Contact-graspnet: Efficient 6-dof grasp generation in cluttered scenes},
  author={Sundermeyer, Martin and Mousavian, Arsalan and Triebel, Rudolph and Fox, Dieter},
  booktitle={2021 IEEE International Conference on Robotics and Automation (ICRA)},
  pages={13438--13444},
  year={2021},
  organization={IEEE}
}

@article{fang2023anygrasp,
  title={Anygrasp: Robust and efficient grasp perception in spatial and temporal domains},
  author={Fang, Hao-Shu and Wang, Chenxi and Fang, Hongjie and Gou, Minghao and Liu, Jirong and Yan, Hengxu and Liu, Wenhai and Xie, Yichen and Lu, Cewu},
  journal={IEEE Transactions on Robotics},
  year={2023},
  publisher={IEEE}
}

@ARTICLE{ycb,
  author={Calli, Berk and Walsman, Aaron and Singh, Arjun and Srinivasa, Siddhartha and Abbeel, Pieter and Dollar, Aaron M.},
  journal={IEEE Robotics \& Automation Magazine}, 
  title={Benchmarking in Manipulation Research: Using the Yale-CMU-Berkeley Object and Model Set}, 
  year={2015},
  volume={22},
  number={3},
  pages={36-52},
  doi={10.1109/MRA.2015.2448951}}

@article{kleeberger2020survey,
  title={A survey on learning-based robotic grasping},
  author={Kleeberger, Kilian and Bormann, Richard and Kraus, Werner and Huber, Marco F},
  journal={Current Robotics Reports},
  volume={1},
  pages={239--249},
  year={2020},
  publisher={Springer}
}

@article{zhang2022robotic,
  title={Robotic grasping from classical to modern: A survey},
  author={Zhang, Hanbo and Tang, Jian and Sun, Shiguang and Lan, Xuguang},
  journal={arXiv preprint arXiv:2202.03631},
  year={2022}
}

@inproceedings{APLoss,
  author       = {J{\'{e}}r{\^{o}}me Revaud and
                  Jon Almaz{\'{a}}n and
                  Rafael S. Rezende and
                  C{\'{e}}sar Roberto de Souza},
  title        = {Learning With Average Precision: Training Image Retrieval With a Listwise
                  Loss},
  booktitle    = {2019 {IEEE/CVF} International Conference on Computer Vision, {ICCV}
                  2019, Seoul, Korea (South), October 27 - November 2, 2019},
  pages        = {5106--5115},
  publisher    = {{IEEE}},
  year         = {2019},
  url          = {https://doi.org/10.1109/ICCV.2019.00521},
  doi          = {10.1109/ICCV.2019.00521},
  bibsource    = {dblp computer science bibliography, https://dblp.org}
}

@article{GraspNet,
  author       = {Chenxi Wang and
                  Haoshu Fang and
                  Minghao Gou and
                  Hongjie Fang and
                  Jin Gao and
                  Cewu Lu},
  title        = {Graspness Discovery in Clutters for Fast and Accurate Grasp Detection},
  journal      = {CoRR},
  volume       = {abs/2406.11142},
  year         = {2024},
  url          = {https://doi.org/10.48550/arXiv.2406.11142},
  doi          = {10.48550/ARXIV.2406.11142},
  eprinttype    = {arXiv},
  eprint       = {2406.11142},
  bibsource    = {dblp computer science bibliography, https://dblp.org}
}

@inproceedings{jiang2025multimodal,
  title={Multimodal llm guided exploration and active mapping using fisher information},
  author={Jiang, Wen and Lei, Boshu and Ashton, Katrina and Daniilidis, Kostas},
  booktitle={Proceedings of the IEEE/CVF International Conference on Computer Vision},
  pages={5392--5404},
  year={2025}
}

@inproceedings{chugao2017oncalibrationofmodernneuralnetworks,
  author       = {Chuan Guo and
                  Geoff Pleiss and
                  Yu Sun and
                  Kilian Q. Weinberger},
  editor       = {Doina Precup and
                  Yee Whye Teh},
  title        = {On Calibration of Modern Neural Networks},
  booktitle    = {Proceedings of the 34th International Conference on Machine Learning,
                  {ICML} 2017, Sydney, NSW, Australia, 6-11 August 2017},
  series       = {Proceedings of Machine Learning Research},
  volume       = {70},
  pages        = {1321--1330},
  publisher    = {{PMLR}},
  year         = {2017},
  url          = {http://proceedings.mlr.press/v70/guo17a.html},
  bibsource    = {dblp computer science bibliography, https://dblp.org}
}

@inproceedings{byeongdo2024equigraspflowse3equivariant6dofgrasppose,
  author       = {Byeongdo Lim and
                  Jongmin Kim and
                  Jihwan Kim and
                  Yonghyeon Lee and
                  Frank C. Park},
  editor       = {Pulkit Agrawal and
                  Oliver Kroemer and
                  Wolfram Burgard},
  title        = {EquiGraspFlow: SE(3)-Equivariant 6-DoF Grasp Pose Generative Flows},
  booktitle    = {Conference on Robot Learning, 6-9 November 2024, Munich, Germany},
  series       = {Proceedings of Machine Learning Research},
  volume       = {270},
  pages        = {5067--5086},
  publisher    = {{PMLR}},
  year         = {2024},
  url          = {https://proceedings.mlr.press/v270/lim25a.html},
  bibsource    = {dblp computer science bibliography, https://dblp.org}
}

@inproceedings{Jonathan2020nonparametericcalibrationforclassification,
  author       = {Jonathan Wenger and
                  Hedvig Kjellstr{\"{o}}m and
                  Rudolph Triebel},
  editor       = {Silvia Chiappa and
                  Roberto Calandra},
  title        = {Non-Parametric Calibration for Classification},
  booktitle    = {The 23rd International Conference on Artificial Intelligence and Statistics,
                  {AISTATS} 2020, 26-28 August 2020, Online [Palermo, Sicily, Italy]},
  series       = {Proceedings of Machine Learning Research},
  volume       = {108},
  pages        = {178--190},
  publisher    = {{PMLR}},
  year         = {2020},
  url          = {http://proceedings.mlr.press/v108/wenger20a.html},
  bibsource    = {dblp computer science bibliography, https://dblp.org}
}

@article{dan2022evaluatingandcalibratinguncertainty,
  author       = {Dan Levi and
                  Liran Gispan and
                  Niv Giladi and
                  Ethan Fetaya},
  title        = {Evaluating and Calibrating Uncertainty Prediction in Regression Tasks},
  journal      = {Sensors},
  volume       = {22},
  number       = {15},
  pages        = {5540},
  year         = {2022},
  url          = {https://doi.org/10.3390/s22155540},
  doi          = {10.3390/S22155540},
  bibsource    = {dblp computer science bibliography, https://dblp.org}
}

@article{DirichletCalib,
  author       = {Meelis Kull and
                  Miquel Perell{\'{o}}{-}Nieto and
                  Markus K{\"{a}}ngsepp and
                  Telmo de Menezes e Silva Filho and
                  Hao Song and
                  Peter A. Flach},
  title        = {Beyond temperature scaling: Obtaining well-calibrated multiclass probabilities
                  with Dirichlet calibration},
  journal      = {CoRR},
  volume       = {abs/1910.12656},
  year         = {2019},
  url          = {http://arxiv.org/abs/1910.12656},
  eprinttype    = {arXiv},
  eprint       = {1910.12656},
  bibsource    = {dblp computer science bibliography, https://dblp.org}
}

@inproceedings{Pavel2019subspaceinferenceforbayesiandeeplearning,
  author       = {Pavel Izmailov and
                  Wesley J. Maddox and
                  Polina Kirichenko and
                  Timur Garipov and
                  Dmitry P. Vetrov and
                  Andrew Gordon Wilson},
  editor       = {Amir Globerson and
                  Ricardo Silva},
  title        = {Subspace Inference for Bayesian Deep Learning},
  booktitle    = {Proceedings of the Thirty-Fifth Conference on Uncertainty in Artificial
                  Intelligence, {UAI} 2019, Tel Aviv, Israel, July 22-25, 2019},
  series       = {Proceedings of Machine Learning Research},
  volume       = {115},
  pages        = {1169--1179},
  publisher    = {{AUAI} Press},
  year         = {2019},
  url          = {http://proceedings.mlr.press/v115/izmailov20a.html},
  bibsource    = {dblp computer science bibliography, https://dblp.org}
}

@article{alireza202confidencecalibrationandpredictiveuncertainty,
  author       = {Alireza Mehrtash and
                  William M. Wells III and
                  Clare M. Tempany and
                  Purang Abolmaesumi and
                  Tina Kapur},
  title        = {Confidence Calibration and Predictive Uncertainty Estimation for Deep
                  Medical Image Segmentation},
  journal      = {{IEEE} Trans. Medical Imaging},
  volume       = {39},
  number       = {12},
  pages        = {3868--3878},
  year         = {2020},
  url          = {https://doi.org/10.1109/TMI.2020.3006437},
  doi          = {10.1109/TMI.2020.3006437},
  bibsource    = {dblp computer science bibliography, https://dblp.org}
}

@inproceedings{gabriel2017regularizingneuralnetwork,
  author       = {Gabriel Pereyra and
                  George Tucker and
                  Jan Chorowski and
                  Lukasz Kaiser and
                  Geoffrey E. Hinton},
  title        = {Regularizing Neural Networks by Penalizing Confident Output Distributions},
  booktitle    = {5th International Conference on Learning Representations, {ICLR} 2017,
                  Toulon, France, April 24-26, 2017, Workshop Track Proceedings},
  publisher    = {OpenReview.net},
  year         = {2017},
  url          = {https://openreview.net/forum?id=HyhbYrGYe},
  bibsource    = {dblp computer science bibliography, https://dblp.org}
}

@inproceedings{seonguk2019learningforsingleshotconfidencecalibration,
  author       = {Seonguk Seo and
                  Paul Hongsuck Seo and
                  Bohyung Han},
  title        = {Learning for Single-Shot Confidence Calibration in Deep Neural Networks
                  Through Stochastic Inferences},
  booktitle    = {{IEEE} Conference on Computer Vision and Pattern Recognition, {CVPR}
                  2019, Long Beach, CA, USA, June 16-20, 2019},
  pages        = {9030--9038},
  publisher    = {Computer Vision Foundation / {IEEE}},
  year         = {2019},
  url          = {http://openaccess.thecvf.com/content\_CVPR\_2019/html/Seo\_Learning\_for\_Single-Shot\_Confidence\_Calibration\_in\_Deep\_Neural\_Networks\_Through\_CVPR\_2019\_paper.html},
  doi          = {10.1109/CVPR.2019.00924},
  bibsource    = {dblp computer science bibliography, https://dblp.org}
}

@inproceedings{Pinto2016Supersizingselfsupervision,
  author       = {Lerrel Pinto and
                  Abhinav Gupta},
  editor       = {Danica Kragic and
                  Antonio Bicchi and
                  Alessandro De Luca},
  title        = {Supersizing self-supervision: Learning to grasp from 50K tries and
                  700 robot hours},
  booktitle    = {2016 {IEEE} International Conference on Robotics and Automation, {ICRA}
                  2016, Stockholm, Sweden, May 16-21, 2016},
  pages        = {3406--3413},
  publisher    = {{IEEE}},
  year         = {2016},
  url          = {https://doi.org/10.1109/ICRA.2016.7487517},
  doi          = {10.1109/ICRA.2016.7487517},
  bibsource    = {dblp computer science bibliography, https://dblp.org}
}

@article{Open3D,
  author       = {Qian{-}Yi Zhou and
                  Jaesik Park and
                  Vladlen Koltun},
  title        = {Open3D: {A} Modern Library for 3D Data Processing},
  journal      = {CoRR},
  volume       = {abs/1801.09847},
  year         = {2018},
  url          = {http://arxiv.org/abs/1801.09847},
  eprinttype    = {arXiv},
  eprint       = {1801.09847},
  bibsource    = {dblp computer science bibliography, https://dblp.org}
}

@inproceedings{DBSCAN,
  author       = {Martin Ester and
                  Hans{-}Peter Kriegel and
                  J{\"{o}}rg Sander and
                  Xiaowei Xu},
  editor       = {Evangelos Simoudis and
                  Jiawei Han and
                  Usama M. Fayyad},
  title        = {A Density-Based Algorithm for Discovering Clusters in Large Spatial
                  Databases with Noise},
  booktitle    = {Proceedings of the Second International Conference on Knowledge Discovery
                  and Data Mining (KDD-96), Portland, Oregon, {USA}},
  pages        = {226--231},
  publisher    = {{AAAI} Press},
  year         = {1996},
  url          = {http://www.aaai.org/Library/KDD/1996/kdd96-037.php},
  bibsource    = {dblp computer science bibliography, https://dblp.org}
}

@article{pi05,
  author       = {Physical Intelligence and
                  Kevin Black and
                  Noah Brown and
                  James Darpinian and
                  Karan Dhabalia and
                  Danny Driess and
                  Adnan Esmail and
                  Michael Equi and
                  Chelsea Finn and
                  Niccolo Fusai and
                  Manuel Y. Galliker and
                  Dibya Ghosh and
                  Lachy Groom and
                  Karol Hausman and
                  Brian Ichter and
                  Szymon Jakubczak and
                  Tim Jones and
                  Liyiming Ke and
                  Devin LeBlanc and
                  Sergey Levine and
                  Adrian Li{-}Bell and
                  Mohith Mothukuri and
                  Suraj Nair and
                  Karl Pertsch and
                  Allen Z. Ren and
                  Lucy Xiaoyang Shi and
                  Laura Smith and
                  Jost Tobias Springenberg and
                  Kyle Stachowicz and
                  James Tanner and
                  Quan Vuong and
                  Homer Walke and
                  Anna Walling and
                  Haohuan Wang and
                  Lili Yu and
                  Ury Zhilinsky},
  title        = {{\(\pi\)}\({}_{\mbox{0.5}}\): a Vision-Language-Action Model with
                  Open-World Generalization},
  journal      = {CoRR},
  volume       = {abs/2504.16054},
  year         = {2025},
  url          = {https://doi.org/10.48550/arXiv.2504.16054},
  doi          = {10.48550/ARXIV.2504.16054},
  eprinttype    = {arXiv},
  eprint       = {2504.16054},
  bibsource    = {dblp computer science bibliography, https://dblp.org}
}
}

\clearpage
\setcounter{page}{1}
\maketitlesupplementary
\section{Derivation for Entropy Definition}
For the conditional entropy $\mathbf{H}[S|G, W]$, we can factorize it as:
\begin{equation}
    \begin{aligned}
        \mathbf{H}[S|G, W] &=  -\int \sum_s p(s, g|w) \log \frac{p(g, s|w)}{p(g|w)} dg \\
                           &=  -\int \sum_s  p(g|w) p(s|g,w) \log p(s|g, w) dg \\
                           &=  -\int   p(g|w) \sum_s \left ( p(s|g,w) \log p(s|g, w) \right ) dg \\
                           &=  \int   p(g|w) \sum_s \left ( -p(s|g,w) \log p(s|g, w) \right ) dg \\
                           &= \mathbb{E}_{p(g|w)} [h(g, w)]
    \end{aligned}
\end{equation}
Here we prove Eq.~\ref{eq:grasp-entropy-define}. For conditional entropy given the observations $\mathcal{D}$, 
\begin{equation}
    \begin{aligned}
        & \mathbf{H}[S|G, W, \mathcal{D}] \\
                           &=  -\int \int \sum_s p(s, g, w|\mathcal{D}) \log \frac{p(s, g, w|\mathcal{D})}{p(g, w|\mathcal{D})} dg dw \\
                           &=  -\int \int \sum_s  p(w|\mathcal{D}) p(g|w) p(s|g,w) \log p(s|g, w, \mathcal{D}) dw dg \\
                           &=  -\int p(w|\mathcal{D}) \int   p(g|w) \sum_s \left ( p(s|g,w) \log p(s|g, w) \right ) dg dw \\
                           &= \mathbb{E}_{p(w|\mathcal{D})} [\eta(w)]
    \end{aligned}
\end{equation}
Here we assume the grasp pose $G$ and observations $G$ are independent given the scene representation $W$.

\section{Derivation for Information Gain}
\label{sec:rationale}
We define the entropy of grasps $\mathbf{H}[g|w]$ as the expectation of the entropy for a single grasp $h(g,w)$, and denote it as a function $\eta(w)$:
\begin{equation}
    \eta(w) = \mathbf{H}[S|G, W] = \mathbb{E}_{p(g|w)} [h(g, w)]
    \label{eq:grasp-entropy-define-append}
\end{equation}
When the scene $w$ is estimated from a set of collected views $\mathcal{D}$, the entropy of the grasp distribution is
\begin{equation}
\begin{aligned}
    \eta(w) &= \mathbb{E}_{p(g,w|\mathcal{D})}[h(g,w)] \\
                                 &= \int p(g,w|\mathcal{D}) h(g,w) dg dw \\
                                 &= \int p(g|w)p(w|\mathcal{D}) h(g,w) dg dw \\
                                 &= \int p(w|\mathcal{D}) \left ( \int p(g|w) h(g,w) \right ) dw \\
                                 &= \int p(w|\mathcal{D}) \eta(w) dw \\
                                 &= \mathbb{E}_{p(w|\mathcal{D})}[\eta(w)]
\end{aligned}
\label{eq:append-entropy}
\end{equation}
From line 2 to line 3, we assume the grasp pose $g$ and the dataset $\mathcal{D}$ are independent when the scene $w$ is given. Using the Taylor expansion of $\eta(w)$ and plugging Eq.~\ref{eq:taylor} into Eq.~\ref{eq:append-entropy}, we get
\begin{equation}
\begin{aligned}
    &\mathbf{E}_{p(w|\mathcal{D})} [\eta(w)] \\ &\approx  \mathbf{E}_{p(w|\mathcal{D})}   [ \eta(w^*) + \nabla_w \eta^T(w^*)(w - w^*) \\ & \qquad \qquad \qquad + \frac{1}{2} (w - w^*)^T \nabla_w^2 \eta(w^*)(w - w^*)  ]
\end{aligned} 
\end{equation}
The expectation of the first term is just itself. The expectation of the second term is 0.
\begin{equation}
\begin{aligned}
    &\mathbf{E}_{p(w|\mathcal{D})}[\nabla_w \eta^T(w^*)(w - w^*)] \\ &= \nabla_w \eta^T(w^*) \mathbf{E}_{p(w|\mathcal{D})}[(w - w^*)] \\ &= \nabla_w \eta^T \mathbf{0} = 0
\end{aligned}
\end{equation}

\begin{table*}[ht]
\centering
\footnotesize
\resizebox{0.85\linewidth}{!}{
\begin{tabular}{lcccccc}
\toprule
 Method & Success & Drop & Fail & Invalid & SR & ECE \\ \hline

ActiveGrasp  ($\lambda = 10^{-1}$)                                            & 309     & 20   & 32   & 39      &  77.25\% &  0.04     \\ 
ActiveGrasp  ($\lambda = 10^{-2}$)                                            & 312     & 20   & 29   & 39      &  78.00\% &  0.04     \\ 
ActiveGrasp  ($\lambda = 10^{-3}$)                                            & 302     & 29   & 32   & 37      &  75.50\% &  0.06     \\ 
ActiveGrasp  ($\lambda = 10^{-4}$)                                            & 316     & 19   & 24   & 41      &  79.00\% &  0.02     \\ 
\bottomrule
\end{tabular}
}
\caption{ \textbf{Ablation study in simulated environments.} We test 4 different settings of $\lambda$ from $10^{-1}$ to $10^{-4}$. \textbf{Drop} is defined as the robot finding the target but failing to lift it. \textbf{Fail} is defined as the robot fails to get in contact with the target, but the grasp model predicts feasible grasps. \textbf{Invalid} is defined as the grasp model that cannot predict any feasible grasps. \textbf{SR} is the success rate. $\textbf{ECE}$ is the Expected Calibration Error computed on all the executed grasps. }
\label{tab:ablation-lambda}
\vspace{-5mm}
\end{table*}

Using $\text{tr}(ABC) = \text{tr}(BCA)$ for any matrices $A, B, C$, the third term is 
\begin{equation}
\begin{aligned}
     &\frac{1}{2} \mathbb{E}_{p(w|\mathcal{D})} \left [  (w - w^*)^T \nabla_w^2 \eta(w^*)(w - w^*) \right ] \\
     &= \frac{1}{2} \mathbb{E}_{p(w|\mathcal{D})} \left [ \text{tr} \left( (w - w^*)^T \nabla_w^2 \eta(w^*)(w - w^*) \right )\right ] \\
     &= \frac{1}{2} \mathbb{E}_{p(w|\mathcal{D})} \left [   \text{tr} \left(  \nabla_w^2 \eta(w^*)(w - w^*)(w - w^*)^T  \right ) \right ] \\
     &= \frac{1}{2}  \text{tr} \left(  \nabla_w^2 \eta(w^*) \mathbb{E}_{p(w|\mathcal{D})} \left [  (w - w^*)(w - w^*)^T \right ]  \right )\\
     &= \frac{1}{2}   \text{tr} \left(  \nabla_w^2 \eta(w^*) \mathbf{H}''[w^*|\mathcal{D}]^{-1}  \right )
\end{aligned}
\end{equation}
We get the results in Eq.~\ref{eq:grasp-entropy}. The mutual information for the candidate view is defined as the difference of the entropy given the candidate $\{x_{\text{acq}}, y_{\text{acq}}\}$:
\begin{equation}
\begin{aligned}
    &\mathbf{I} [g, w; y_{\text{acq}} | x_{\text{acq}}, \mathcal{D}] \\ &= \mathbf{H}[g, w|D]  - \mathbf{H}[g, w |\{x_{\text{acq}},  y_{\text{acq}}\}, \mathcal{D}] \\
        &= \mathbf{E}_{p(w|\mathcal{D})} [\eta(w)] - \mathbf{E}_{p(w|x_{\text{acq}}, y_{\text{acq}}, \mathcal{D})} [\eta(w)]
\end{aligned}
\end{equation}
Using the Gaussian approximation, $p(w|\mathcal{D}) \sim \mathcal{N}(w^*, \mathbf{H}''[w^*|\mathcal{D}]^{-1})$ and $p(w|x_{\text{acq}},  y_{\text{acq}}\mathcal{D}) \sim \mathcal{N}(\tilde{w}^*, \mathbf{H}''[\tilde{w}^* | x_{\text{acq}},  y_{\text{acq}}, \mathcal{D}]^{-1})$, where $\tilde{w}^*$ is the new MAP estimate. We expand $\eta(w)$ near the previous MAP estimate $w^*$ instead of the new MAP estimate $\tilde{w}^*$, 
\begin{equation}
\begin{aligned}
    &\mathbf{E}_{p(w|x_{\text{acq}}, y_{\text{acq}}, \mathcal{D})} [\eta(w)] \\ &\approx  \mathbf{E}_{p(w|x_{\text{acq}}, y_{\text{acq}}, \mathcal{D})}[\eta(w^*) + \\ & \qquad \nabla_w \eta^T(w^*)(w - w^*) + \frac{1}{2} (w - w^*)^T \nabla_w^2 \eta(w^*)(w - w^*)]
\end{aligned}
\end{equation}
The expectation is:
\begin{equation}
\begin{aligned}
     &\mathbf{E}_{p(w|x_{\text{acq}}, y_{\text{acq}}, \mathcal{D})} [\eta(w)]  \\
            &\approx \eta(w^*) + \nabla \eta(w^*)^T (\tilde{w}^* - w^*) \\ & \qquad + \frac{1}{2}(\tilde{w}^* - w^*)^T \nabla^2\eta(w^*)(\tilde{w}^* - w^*) \\ & \qquad \qquad + \frac{1}{2} \text{tr} \left ( \nabla^2\eta(w^*) \mathbf{H}''(w^*|x_{\text{acq}}, y_{\text{acq}}, \mathcal{D})^{-1} \right )
\end{aligned}
\end{equation}
Plugging back into the mutual information gain formula, we get:
\begin{equation}
    \begin{aligned}
        &\mathbf{I} [G,W; y_{\text{acq}} | x_{\text{acq}}, \mathcal{D}]  \\&\approx   \frac{1}{2}  \text{tr}  \left ( \nabla_w^2  \eta(w^*) \left [ \mathbf{H}''[w^*|\mathcal{D}]^{-1} - \mathbf{H}''[\tilde{w}^*|x_{\text{acq}}, y_{\text{acq}}, \mathcal{D}]^{-1} \right ] \right ) \\ 
            & \qquad - \nabla \eta(w^*)^T (\tilde{w}^* - w^*) - \frac{1}{2}(\tilde{w}^* - w^*)^T \nabla^2\eta(w^*)(\tilde{w}^* - w^*) 
    \end{aligned}
\end{equation}
If we assume the shift from $w^*$ to the new estimate $\tilde{w}^*$ is small, then the mutual information is approximated as 
\begin{equation}
    \begin{aligned}
        &\mathbf{I} [g,w; y_{\text{acq}} | x_{\text{acq}}, \mathcal{D}]  \\ & \approx   \frac{1}{2}  \text{tr}  \left ( \nabla_w^2  \eta(w^*) \left [ \mathbf{H}''[w^*|\mathcal{D}]^{-1} - \mathbf{H}''[w^*|x_{\text{acq}}, y_{\text{acq}}, \mathcal{D}]^{-1} \right ] \right )
    \end{aligned}
\end{equation}
Since the computation of $\mathbf{H}''$ (Eq.~\ref{eq:Hessian-compute}) only involves $x$, we can omit $y_{\text{acq}}$ in the equation and get 
\begin{equation}
    \begin{aligned}
        &\mathbf{I} [G,W; y_{\text{acq}} | x_{\text{acq}}, \mathcal{D}]  \\ & \approx  \frac{1}{2}  \text{tr}  \left ( \nabla_w^2  \eta(w^*) \left [ \mathbf{H}''[w^*|\mathcal{D}]^{-1} - \mathbf{H}''[w^*|x_{\text{acq}},  \mathcal{D}]^{-1} \right ] \right )
    \end{aligned}
\end{equation}
Here we get the results in Eq.~\ref{eq:mutual-information}.

\section{Derivation for $\nabla_w\eta(w)$}
$\eta(w)$ is defined as:
\begin{equation}
    \eta(w) = \mathbb{E}_{p(g|w)} [h(g, w)]=\int p(g|w) h(g,w) dg
\end{equation}
Taking the first-order derivative w.r.t $w$, we have
\begin{equation}
    \nabla_w \eta(w) = \int (\nabla_w h(g,w))p(g,w) + h(g,w) \nabla_w p(g, w) dg
    \label{eq:append-first-order-pre}
\end{equation}
Using the equation
\begin{equation}
    \nabla_w p(g|w) = p(g|w) \frac{\nabla_w p(g|w)}{p(g|w)} = p(g|w ) \nabla_w\log p(g|w)
\end{equation}
Plugging in, we get the following.
\begin{equation}
\begin{aligned}
    &\nabla_w \eta(w) \\ &= \int p(g|w) \left (\nabla_w h(g,w) + h(g,w) \nabla_w \log p(g| w) \right) dg \\
                     &= \mathbb{E}_{p(g|w)}[\nabla_w h(g,w) + h(g,w) \nabla_w \log p(g|w)]
\end{aligned}
\end{equation}
Since we model the grasp distribution using an energy-based model, we have
\begin{equation}
   \log p(g|w) = -E_\theta(g, w, \sigma_k) + \log Z; \; Z= \int e^{-E_\theta(g, w, \sigma_k)} dg
\end{equation}
Take the first-order derivative,
\begin{equation}
   \nabla_w \log p(g|w) = - \nabla_w E_\theta(g, w, \sigma_k) + \frac{1}{Z} \nabla_w Z
\end{equation}
The first-order derivative of $Z$ is 
\begin{equation}
\begin{aligned}
    \frac{1}{Z} \nabla_w Z &= \frac{1}{Z} \nabla_w \int e^{-E_\theta(g, w, \sigma_k)} dg \\
                           &= \frac{1}{Z} \int e^{-E_\theta(g, w, \sigma_k)} (- \nabla_wE_\theta(g, w, \sigma_k)) dg \\
                           &=\int p(g|w) (- \nabla_wE_\theta(g, w, \sigma_k)) dg \\
                           &= -\mathbb{E}_{p(g|w)}[\nabla_wE_\theta(g, w, \sigma_k)]
\end{aligned}
\end{equation}
From line 2 to line 3, we put $Z$ inside the integral and use the definition in Eq.~\ref{eq:ebm-definition}. Putting this back in Eq.~\ref{eq:append-first-order-pre}, we get
\begin{equation}
\begin{aligned}
    \nabla_w \eta(w) &= \int p(g|w) \left (\nabla_w h(g,w) + h(g,w) \nabla_w \log p(g| w) \right) dg \\
                     &= \mathbb{E}_{p(g|w)}[\nabla_w h(g,w) + \\
                     & \qquad h(g,w) ( - \nabla_w E_\theta(g, w, \sigma_k) - \\
                     & \qquad \qquad \mathbb{E}_{p(g|w)}[\nabla_wE_\theta(g, w, \sigma_k)])] \\
                     &= \mathbb{E}_{p(g|w)}[\nabla_w h(g,w)] - \\
                     & \qquad \mathbb{E}_{p(g|w)}[h(g,w)\nabla_w E_\theta(g, w, \sigma_k]  - \\
                     & \qquad \qquad \mathbb{E}_{p(g|w)}[h(g,w)]\mathbb{E}_{p(g|w)}[\nabla_wE_\theta(g, w, \sigma_k)])]
\end{aligned}
\end{equation}
Here we prove Eq.~\ref{eq:eta-first-order}. We sample from the energy-based model to compute the expectation.

\begin{figure}[ht]
    \centering
    \includegraphics[width=\linewidth]{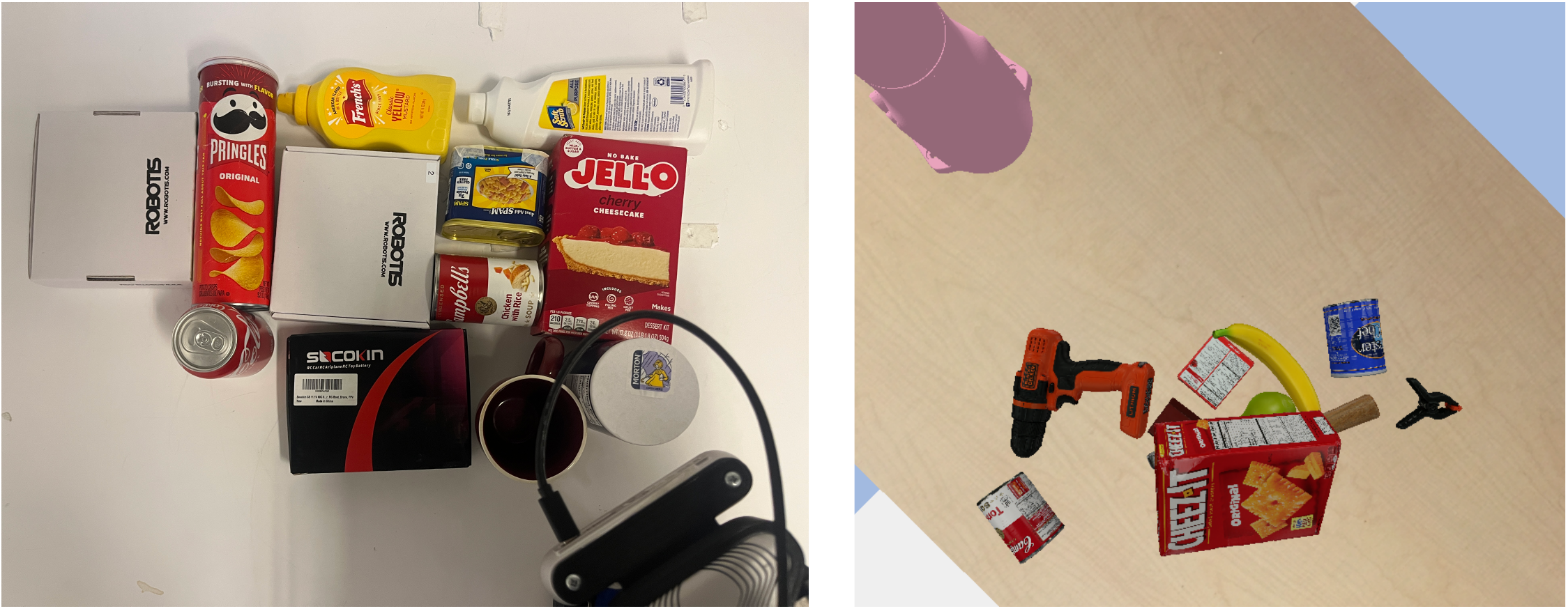}
    \caption{\textbf{Visualization for Real World and Simulation Experiment Setup} We show the objects used in the real world experiment~(left) and one randomly generated simulation scene~(right).}
    \label{fig:setup}
    \vspace{-5mm}
\end{figure}

\begin{table*}[!ht]
\centering
\resizebox{0.98\textwidth}{!}{
\begin{tabular}{lccccccc}
\toprule
              & AP $\uparrow$  &  ECE $\downarrow$  & Ang (acc.) $\downarrow$  & Trans (acc.) $\downarrow$  & Ang (rec.) $\downarrow$  & Trans (rec.) $\downarrow$  & ECE (Bullet) $\downarrow$ \\ \hline 
Se3diff+Scene+FSM(normalized)+AP+T & 81.52  & 0.03 & 0.3900 & 0.0242 & 0.3294 & 0.0203 & 0.06    \\ 
Se3diff+Scene+FSM(unnormalized)+AP+T & 87.84  & 0.03 & 0.3764 & 0.0212 & 0.3068 & 0.0178 & 0.05    \\ 
\bottomrule
\end{tabular}
}
\caption{ \textbf{Ablation study of energy-based model on the ACRONYM dataset~\cite{ACRONYM}}. \textbf{AP}: average precision using the predicted energy. \textbf{ECE}: Expected Calibration Error on Acronym dataset. \textbf{Ang/Trans (acc.)}: Average closest angular and translational distance from the closest ground-truth successful grasps to the predicted grasps. \textbf{Ang/Trans(rec.)}: Average closest angular and translational distance from the predicted grasps to the closest ground-truth successful grasps. The unit of angular and translational distance is expressed as radians and meters. \textbf{ECE (Bullet)}: Expected Calibration Error after executing grasps in the Bullet simulator. \textbf{Scene}: Our implementation of the model augmented with scene information. \textbf{FSM(normalized)}: Model trained with both success and failure grasp, and the probability is normalized. \textbf{FSM(unnormalized)}: Model trained with both success and failure grasp, and the probability is unnormalized. \textbf{AP} Model trained with AP Loss. \textbf{T}: Model trained with learnable temperature. }
\label{tab:normalization-ablation}
\vspace{-5mm}
\end{table*}

\section{Experiment Details}
To initialize the 3DGS, we unproject the RGBD image to the 3D space and use it to initialize the 3DGS.
We use the forward kinematics of the robot arm to get the camera pose for each captured view to train the 3DGS. 
Each Gaussian is set to isotropic. 
The Spherical Harmonics degree is set to 0.
We train the 3DGS for 1k steps after we add one new view.
The pruning and duplicating process runs every 500 steps.
To compute the expectation of information gain, we sample 512 grasp poses from the energy-based model.
We also show the experiment setup for both simulation and real-world experiments in Fig.~\ref{fig:setup}. 

\section{Regularization Parameter $\lambda$}
We do an ablation study on the regularization constant $\lambda$. We follow the same setting as in the simulation experiment in Sec~\ref {sec:sim-result}. Each setting starts with 2 fixed initial views and actively takes 2 more views before grasping. The result is summarized in Tab.~\ref{tab:ablation-lambda}. $\lambda = 10^{-4}$ achieves the best result,  a success rate of 79.00\% as we report in the paper in Tab.~\ref{tab:sim-results}. Even the worst case ($\lambda=10^{-3}$) has a higher success rate than other active view selection methods ($74.25\%$), when the number of selected views is the same. 

\begin{table}[ht]
    \centering
    \begin{tabular}{lc}
    \toprule
    Method &  Success Rate \\
    \midrule
    ActiveGrasp (N=128) & 74.50\%  \\
    ActiveGrasp (N=256) & 75.25\%  \\
    ActiveGrasp (N=512) & 79.00\%  \\
    \bottomrule
    \end{tabular}
    \captionof{table}{\textbf{Number of grasp poses for information gain estimation} $N$ is the number of grasp poses used to compute in Eq.~\ref{eq:eta-first-order}. }
    \label{tab:ablation-number-grasp-pose}
    \vspace{-5mm}
\end{table}

\section{Ablation on Information Gain Estimation}
\begin{figure}[t]
    \centering
    \includegraphics[width=\linewidth]{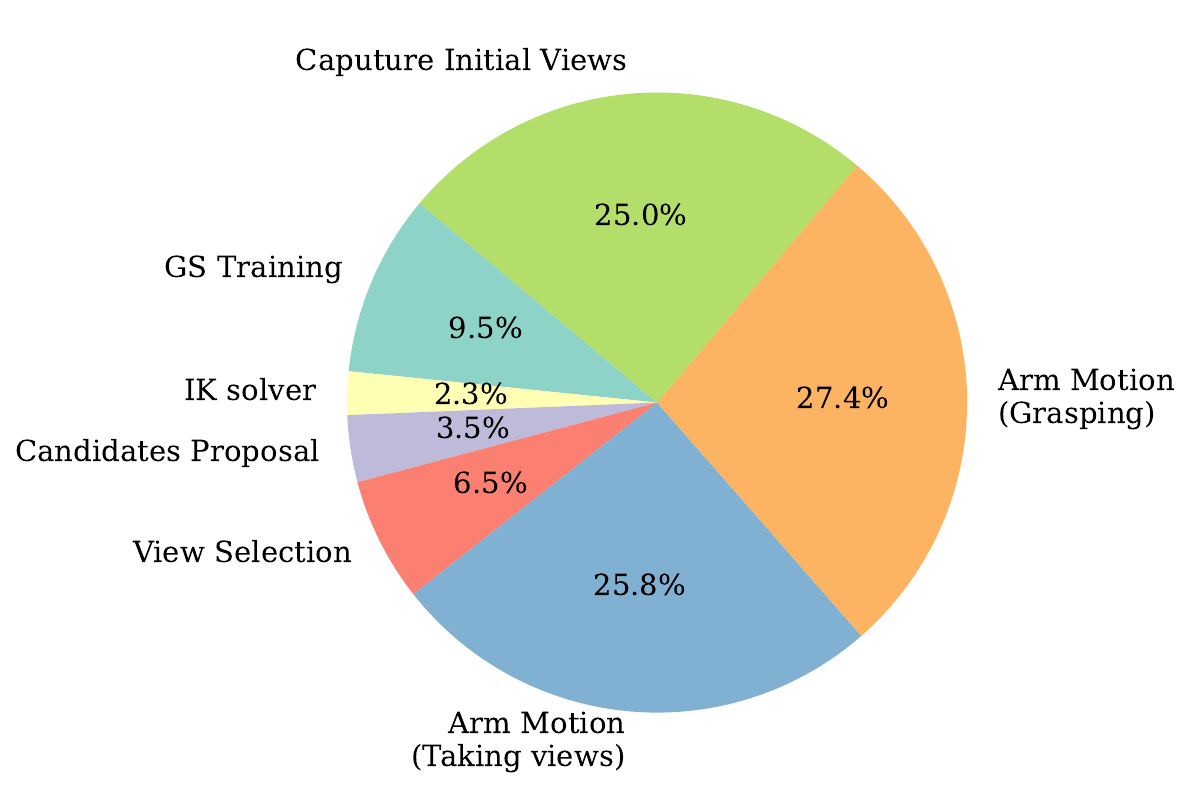}
    \caption{\textbf{Runtime Analysis of our Active Grasping System.} 
    As we can see, the view selection module and 3D-GS data representation do not create significant overhead.}
    \label{fig:time-analysis}
    \vspace{-5mm}
\end{figure}

We do an ablation study on the number of grasp poses used to estimate the information gain. We follow the same setting as in the simulation experiment in Sec~\ref {sec:sim-result}. The result is summarized in Tab.~\ref{tab:ablation-number-grasp-pose}. As the number of grasp poses increases, the estimation becomes better, leading to better performance. We use 512 grasp poses in the main paper. As shown in Fig.~\ref{fig:time-analysis}, even though we use 512 grasp poses, the view selection process only accounts for 6.5\% of the entire running time. 

\section{Network Parameterization}
We formulate the predicted probability as unnormalized in Eq.~\ref{eq:dirich-prob} because we find it is better than a normalized formulation as:
\begin{equation}
    p_S = \frac{e^{a_S} + 1}{e^{a_S} + e^{a_F} + 2}; \quad p_F = \frac{e^{a_F} + 1}{e^{a_S} + e^{a_F} + 2}
\end{equation}
We carry out an ablation study, and the result is summarized in Tab.~\ref{tab:normalization-ablation}. As shown in the table, the unnormalized version has a lower Ang/Trans(acc.) and Ang/Trans(rec.) error than the normalized version, showing that the generation quality is better. We think the reason is that the unnormalized formulation allows the network to predict a grasp pose as neither success nor failure. Therefore, it gives the network more freedom to adjust the energy landscape. 


\section{Time Analysis}
We provide a runtime analysis of our active grasping system in Fig~\ref{fig:time-analysis}. 
The results show that the view selection module and 3D-GS data representation do not create significant overhead, and the overall runtime is efficient for real-world applications, as most of the time is spent on the motion of the robot arm.

\section{Comparison on GraspNet-1billion}
We evaluated our method using the GraspNet-1Billion dataset following the same setup in ActiveNGF~\cite{ma2024activeNGF}.
We use the same parameter setting and select 10 views from each scene.
Then we use the same grasp predictor from ActiveNGF to predict the poses and compute the metric.
The result is summarized in Table.~\ref{tab:GS1B-result}.

\begin{table}[ht]
\centering

\resizebox{.5\textwidth}{!}{
    \begin{tabular}{l|ccc|ccc|ccc}
    \hline
    \multirow{2}{*}{Methods} 
    & \multicolumn{3}{c|}{Seen} 
    & \multicolumn{3}{c|}{Similar} 
    & \multicolumn{3}{c}{Novel} \\
    & AP & AP$_{0.8}$ & AP$_{0.4}$ 
    & AP & AP$_{0.8}$ & AP$_{0.4}$ 
    & AP & AP$_{0.8}$ & AP$_{0.4}$ \\
    \hline
    ActiveNGF 
    & 55.12 & 65.07 & 48.88 
    & 52.85 & 62.63 & 46.49 
    & 24.74 & 30.21 & 12.00 \\
    Ours & 58.44  &  66.33 &   51.67
    &  56.22  &  65.67 & 48.33  
    & 26.44 & 35.33 & 14.00 \\
    \hline
    \end{tabular}
}
\caption{\textbf{Graspnet-1Billion Experiment}}
\label{tab:GS1B-result}
\vspace{-8mm}
\end{table}

\end{document}